\documentclass[usenames,dvipsnames]{zgroup}
\usepackage[utf8]{inputenc}

\usepackage{times}
\usepackage{hyperref}
\usepackage{url}

\usepackage{amsmath,amsfonts,bm}

\def\eqref#1{equation~\ref{#1}}

\def\1{\bm{1}}

\DeclareMathAlphabet{\mathsfit}{\encodingdefault}{\sfdefault}{m}{sl}
\SetMathAlphabet{\mathsfit}{bold}{\encodingdefault}{\sfdefault}{bx}{n}

\usepackage{hyperref}
\usepackage{url}

\usepackage{booktabs}
\usepackage{xcolor}
\usepackage{amssymb}
\usepackage{pifont}
\usepackage{dsfont}
\usepackage{graphicx}
\usepackage{natbib}
\usepackage{graphicx}
\usepackage{appendix}

\usepackage{mathtools} 
\usepackage{booktabs} 
\usepackage{tikz} 
\usepackage{lipsum}
\usepackage{algorithm}
\usepackage{algpseudocode}
\usepackage[bb=px]{mathalpha}
\usepackage{xspace}
\usepackage{enumitem}
\usepackage{mdframed}
\usepackage{yhmath}
\usepackage{wrapfig}

\newcommand{\methodName}{SpAR\xspace}
\newcommand{\methodNameFull}{Spectral Adapted Regressor\xspace}
\newtheorem{proposition}{Proposition}
\newtheorem{lemma}{Lemma}
\newtheorem{corollary}{Corollary}[theorem]
\newcommand{\Var}{\mathrm{Var}}
\newcommand{\Bias}{\mathrm{Bias}}
\newcommand{\wproj}{w_{\mathrm{proj}}}
\newcommand{\Rows}{\mathrm{Rows}}
\newcommand{\Span}{\mathrm{Span}}
\newcommand{\T}{\top}
\newcommand{\CDF}{\mathrm{CDF}}

\usepackage[utf8]{inputenc} 
\usepackage[T1]{fontenc}    
\usepackage{hyperref}       
\usepackage{url}            
\usepackage{booktabs}       
\usepackage{amsfonts}       
\usepackage{nicefrac}       
\usepackage{microtype}      
\usepackage{xcolor}         
\usepackage{bbm}

\title[Out of the Ordinary: Spectrally Adapting Regression for Covariate Shift]{Out of the Ordinary: Spectrally Adapting \\ Regression for Covariate Shift}
\author{\Name{Benjamin Eyre$^1$}
\Email{beneyre@cs.columbia.edu}\\
\Name{Elliot Creager$^{2,3}$}
\Email{creager@uwaterloo.ca}\\
\Name{David Madras$^{4}$}
\Email{dmadras@google.com}\\
\Name{Vardan Papyan$^{2,5}$}
\Email{vardan.papyan@utoronto.ca}\\
\Name{Richard Zemel$^{1}$}
\Email{zemel@cs.columbia.edu}\\
\addr $^1$Columbia University $^2$ University of Waterloo $^3$Vector Institute $^4$ Google Research $^5$University of Toronto}
\date{February 2021}

\begin{document}

\setlength{\belowdisplayskip}{7pt} \setlength{\belowdisplayshortskip}{7pt}
\setlength{\abovedisplayskip}{7pt} \setlength{\abovedisplayshortskip}{7pt}

\widowpenalty-1000

\maketitle

\begin{abstract}
Designing deep neural network classifiers that perform robustly on distributions differing from the available training data is an active area of machine learning research.
However, out-of-distribution generalization for regression---the analogous problem for modeling continuous targets---remains relatively unexplored.
To tackle this problem, we return to first principles and analyze how the closed-form solution for Ordinary Least Squares (OLS) regression is sensitive to covariate shift.
We characterize the out-of-distribution risk of the OLS model in terms of the eigenspectrum decomposition of the source and target data.
We then use this insight to propose a method for adapting the weights of the last layer of a pre-trained neural regression model to perform better on input data originating from a different distribution. We demonstrate how this lightweight spectral adaptation procedure can improve out-of-distribution performance for synthetic and real-world datasets.
\end{abstract}

\section{Introduction}

Despite their groundbreaking benchmark performance on many tasks---from image recognition and natural language understanding to disease detection \citep{balagopalan2020cross, krizhevsky2017imagenet, devlin2019bert}---deep neural networks (DNNs) tend to underperform when confronted with data that is dissimilar to their training data \citep{geirhos2020shortcut,d2020underspecification,arjovsky2019invariant,koh2021wilds}.  

Understanding and addressing \emph{distribution shift} is critical for the real-world deployment of machine learning (ML) systems.
For instance, 
datasets from the WILDS benchmark
\citep{koh2021wilds} provide real-world case studies suggesting that poor performance at the subpopulation level can have dire consequences in crucial applications such as monitoring toxicity of online discussions, or tumor detection from medical images.
Furthermore, \citet{degrave2021ai} demonstrated that models trained to detect COVID-19 from chest X-Rays performed worse when evaluated on data gathered from hospitals that were not represented in the training distribution.
Unfortunately, poor out-of-distribution (OOD) generalization remains a key obstacle to broadly deploying ML models in a safe and reliable way.

While work towards remedying these OOD performance issues has been focused on classification, predicting 
continuous targets under distribution shift has received less attention. 
In this paper, we present a lightweight method for updating the weights of a pre-trained regression model (typically a neural network, in which case only the final layer is updated).
This method is motivated by a theoretical analysis that yields a concrete reason,
which we call {\it Spectral Inflation}, to explain why regressors may fail under covariate shift, a specific form of distribution shift.
We then propose a post-processing method
that improves the OOD performance of regression models in a synthetic experiment and several real-world datasets.

\section{Background}
Distribution shift problems involve training on inputs $X$ and target labels $Y$ sampled from $P(X,Y)$, then evaluating the resulting model on a distinct distribution $Q(X,Y)$.
Several learning frameworks consider different
forms of distribution shift, depending on the structure of $P$ and the degree of prior knowledge about $Q$ that is available.
For example in Domain Adaptation (DA)~\citep{ben2006analysis}, unlabelled data ({\it unsupervised DA})  or a small number of labelled examples ({\it semi-supervised DA}) from $Q$ are used to adapt a model originally trained on samples from $P$. In some of our experiments, we conduct unsupervised DA. In others 
the setting is very similar to unsupervised DA, with the exception that 
we update our model directly on the unlabeled test examples $X_{te} \sim Q(X)$ rather than on an independent sample $X' \sim Q(X)$ not used for evaluation.
This setting is realistic and relevant to machine learning
\citep{shocher2018zero, sun2020test, bau2020semantic}.

We also assume the distribution shift is due to \emph{covariate shift},
where the conditional distribution over the evaluation data $Q(Y|X)$ is equal to the conditional distribution over the training data $P(Y|X)$, but the input marginals $P(X)$ and $Q(X)$ differ.
This broadly studied assumption \citep{sugiyama2007covariate,gretton2009covariate,ruan2022optimal} states that the sample will have the same relationship to the label in both distributions.
Within this setting, we turn our attention to the regression problem.

\section{Robust Regression by Spectral Adaptation}
Least-squares regression has a known closed-form solution that minimizes the training loss, and yet this solution is not robust to covariate shift.
In this section we show \emph{why} this is the case by characterizing the OOD risk in terms of 
the eigenspectrum 
of the source and (distribution-shifted) target data.
We then use insights from our theoretical analysis to derive a practical post-processing algorithm that uses unlabeled target data to adapt the weights of a regressor previously pre-trained on labeled source data.
The adaptation is done in the 
spectral
domain by first identifying subspaces of the target and source data that are misaligned, then projecting out the pre-trained regressor's components along these subspaces.
We call  our method \textbf{\methodNameFull (\methodName)}.

\begin{figure}[!t]
  \centering
  \includegraphics[width=1\linewidth]{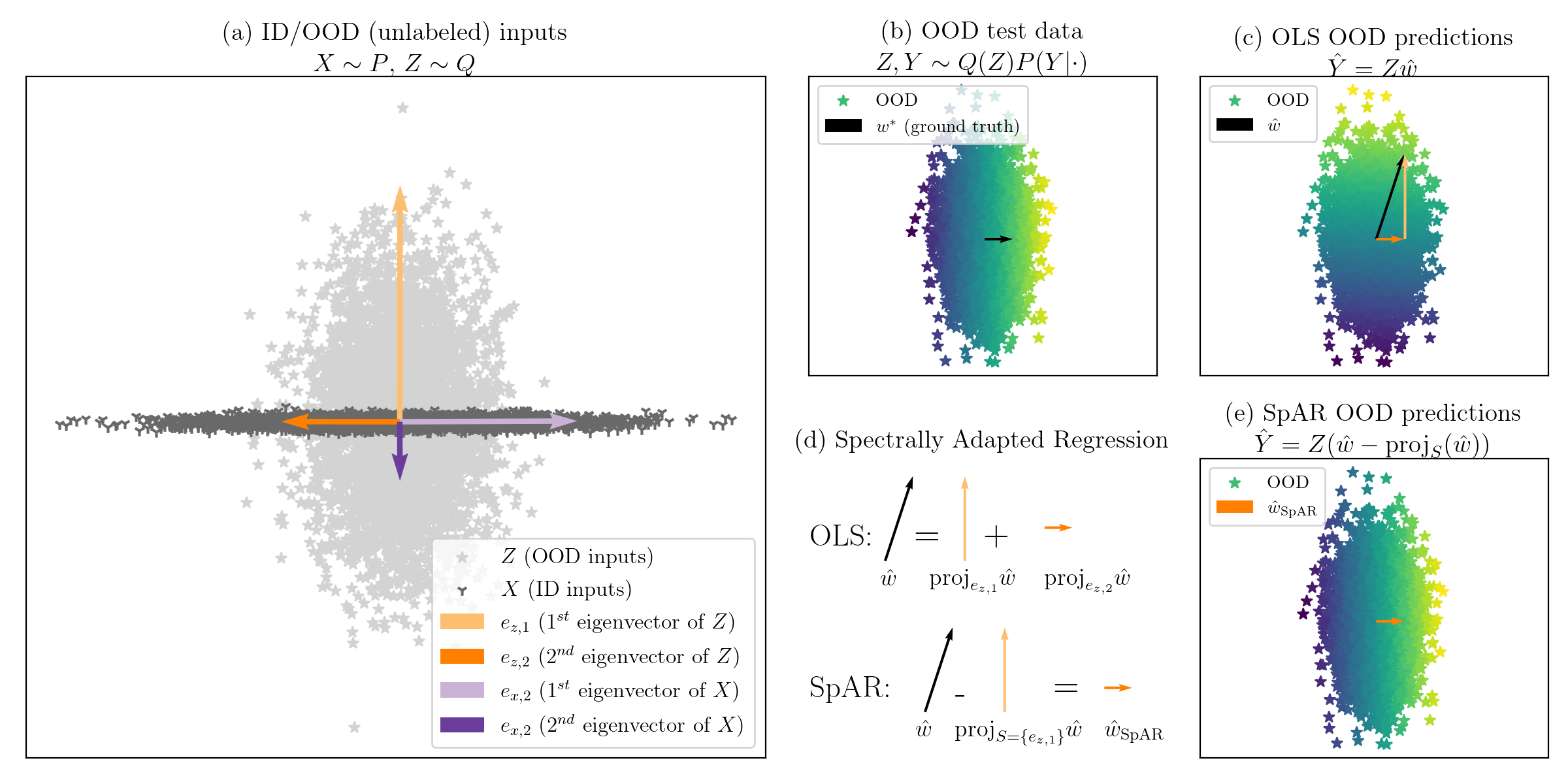}
  \caption{\textbf{Ordinary Least Squares Regression under Covariate Shift.}
  (a) Points are 2D input samples in the training set $X$ and test set $Z$. The in-distribution (ID) training data demonstrates nearly zero vertical variance, while the out-of-distribution (OOD) test data varies significantly in this direction. (b) Samples in $Z$ colored 
  according to their true, noiseless labels $Zw^*$. (c) Samples in $Z$ colored according to their OLS predictions $Z\hat{w}$.  
  Crucially, to minimize training risk, \emph{OLS learns to weigh the vertical component highly} causing erroneous predictions OOD.
  (e) 
  \methodName identifies a spectral subspace $S$ where train/test variance differ the most, and projects it out.
  Thus, the regressor created by \methodName ignores the direction with 
  high variance
  and nearly recovers $w^*$.
  }\label{fig:labels_vs_predictions}
  
\end{figure}

\subsection{Analyzing OLS Regression Under Covariate Shift}\label{sec:data_generative}

We begin with the standard Ordinary Least Squares (OLS) data generating process \citep{murphy2022probabilistic}. Rows of the input data matrix, $X \in \mathbb{R}^{N\times D}$, are i.i.d. samples from an unknown distribution $P$ over $\mathbb{R}^D$; these can be any representation, including one learned by a DNN from training samples. The rows of the evaluation input data, $Z \in \mathbb{R}^{M\times D}$, are generated using a different distribution $Q$ over $\mathbb{R}^D$. Analyzing final layer representations is useful as DNN architectures typically apply linear models to these to make predictions.  Targets depend on $X$ and $w^*$, a labeling vector in $\mathbb{R}^D$, and a noise term $\epsilon$. The targets associated with the test
data $Z$ use the same true labeling vector $w^*$ but do not include a noise term as it introduces irreducible error:

\begin{equation}\label{eq:x_data_generation}
    X\sim P^N, \quad Y_X = Xw^* + \epsilon, \quad \epsilon\sim \mathcal{N}(0, \sigma^2I), \quad Z\sim Q^M, \quad  Y_Z = Zw^*.
\end{equation}

The estimated regressor $\hat{w}$ that minimizes the expected squared error loss has the following form \citep{murphy2022probabilistic}, using $X^{\dagger}$, the Moore-Penrose Pseudoinverse of $X$, and its singular value decomposition, $X^{\dagger} = V_XD_X^{\dagger}U_X^\T$:
\begin{equation}\label{eq:pseudoinv_soln}
    \arg\min_{w}\mathbb{E}[\|Y_X - Xw\|^2_2] = \hat{w} = X^{\dagger}Y_X = V_XD_X^{\dagger}U_X^\T Y_X.
\end{equation}
We refer to $\hat w$ as the ``OLS regressor'' or ``pseudoinverse solution''.
Our primary expression of interest will be the expected loss of $\hat w$ under covariate shift, which is the squared error between the true labels $Y_Z$ and the values predicted by our estimator $\hat{w}$. Specifically, we will analyze the expression:
\begin{equation}
    \mathrm{Risk}_{\mathrm{OLS-OOD}}(\hat w) = \mathbb{E}[\|Y_Z - Z\hat{w}\|^2_2].
\end{equation}

In addition to using the singular value decomposition $X = U_XS_XV_X^\top$, we can also use the singular value decomposition of the target data $Z = U_ZS_ZV_Z^\top$. 
We define $\lambda_{x,i}, \lambda_{z,i}$ to be the $i^{th}$ singular values of $X$ and $Z$, respectively, and $e_{x,i}, e_{z,i}$ their corresponding unit-length right singular vectors. We will also refer to $\lambda_{x,i}^2, \lambda_{z,i}^2$ and $e_{x,i}, e_{z,i}$ as eigenvalues/eigenvectors, as they comprise the eigenspectrum of the uncentered covariance matrices $X^\top X$ and $Z^\top Z$. We use the operator $\Rows()$ to represent the set containing the rows of a matrix. The OOD risk of $\hat{w}$ is presented in the following theorem in terms of interaction between the eigenspectra of $X$ and $Z$:

\begin{theorem}\label{thm:ols_loss}
Assuming the data generative procedure defined in Equations \ref{eq:x_data_generation}, and that $w^* \in \Span(\Rows(X))$ and $\Rows(Z) \subset \Span(\Rows(X))$, the OOD squared error loss of the estimator $\hat{w} = X^{\dagger}Y$ is equal to:
\begin{equation}\label{eq:variance}
    \mathbb{E}[\|Y_Z - Z\hat{w}\|^2_2] = \sigma^2 \sum_{i=1}^D\sum_{j = 1}^D\frac{\lambda_{z,j}^2}{\lambda_{x,i}^2} \langle e_{x,i}, e_{z,j}\rangle^2 \mathbbm{1}[\lambda_{x,i} > 0].
\end{equation}
\end{theorem}
 
\begin{figure}[!t]
  \centering
  \includegraphics[width=0.75\linewidth]{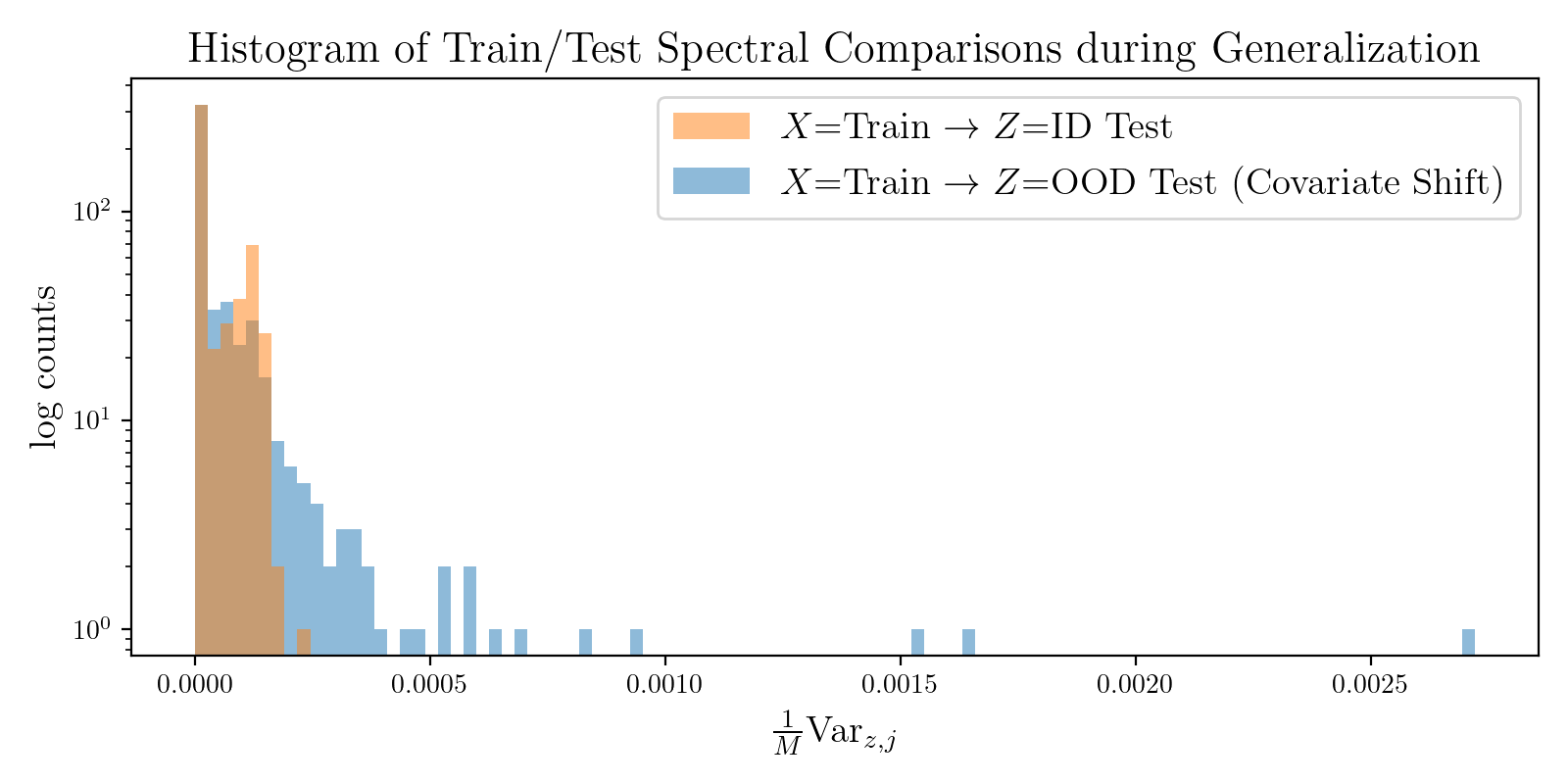}
  \caption{
  \textbf{Spectral Inflation.}
  We use the PovertyMap-WILDS dataset~\citep{koh2021wilds} to investigate how input spectra change when a regressor trained on real-world data generalizes to (perhaps shifted) test data.
  $X$ and $Z$ are composed of representations from a DNN.
  $Z$ represents data either from an in-distribution or out-of-distribution test set.
  $\Var_{z,j}$, as defined in Equation \ref{eq:individual_loss_contributions}, measures the amount of \emph{Spectral Inflation}---small amounts of training set variation becoming large at test time---occurring along a given test eigenvector.
  Because each test sample has a different number of examples $M$, we normalize for a fair comparison.
  We see that when $Z$ is an out-of-distribution sample, much more spectral inflation occurs than when we generalize to an in-distribution sample.
  }\label{fig:spectral_inflation}
  
\end{figure}

This theorem indicates that if the samples in $Z$ present a large amount of variance along the vector $e_{z,j}$, resulting in a large eigenvalue $\lambda_{z,j}^2$,  but the training set $X$ displays very little variance along vectors at very similar angles, $\hat{w}$ will incur high loss. We refer to this scenario, when an eigenvector demonstrates this spike in variance at test time, as
\textbf{Spectral Inflation}.
An illustration of Spectral Inflation and its consequences are depicted in Figure \ref{fig:labels_vs_predictions}, and we present evidence of Spectral Inflation occurring in DNN representations in a real-world dataset in Figure \ref{fig:spectral_inflation}.
The analysis follows from the cyclic property of the trace operator, which allows us to isolate the noise term $\epsilon$.
This, in turn, enables a decomposition of the remaining expression in terms of the two eigenspectra of $Z^\T Z$ and $X^\T X$.
A full derivation of this decomposition is available in Appendix \ref{sec:appendix_ols_covariate}.

\subsection{Spectral Adaptation Through Projection}\label{sec:adaptation_method}

We now focus on
identifying the eigenvectors occupying the rows of $V_Z^\top$ that contribute 
significantly
to the expected loss described in Equation \ref{eq:variance}, and use them to construct a subset $S \subseteq \Rows(V_Z^\top)$. We then use $S$ to construct a new regressor $\wproj$, by projecting $\hat{w}$ onto the subspace spanned by the eigenvectors in $S^c$, the complement of $S$:

\begin{equation}\label{eq:projected_regressor}
    \wproj = \hat{w} - \sum_{e \in S}\langle \hat{w}, e\rangle e.
\end{equation}

This regressor is not influenced by the Spectral Inflation displayed along each eigenvector in $S$, 
as $\wproj$ exists in a subspace orthogonal to the subspace spanned by the vectors in $S$. Specifically, we can decompose the loss for this estimator $\wproj$ into a sum over each eigenvector in $\Rows(V_Z^\top)$, where the contribution of the eigenvector $e_{z,j}$ to the loss is determined by whether that eigenvector is included in the set $S$. The following theorem expresses the expected OOD loss of $\wproj$: 

\begin{theorem}\label{thm:bias_variance_decomp}
Taking on the same assumptions as Theorem \ref{thm:ols_loss}, the regressor $\wproj$ constructed using a set $S \subseteq \Rows(V_Z^\top)$ as defined in
Equation \ref{eq:projected_regressor}, has the following expected OOD squared error loss:
\begin{align}
     & \mathbb{E}[\|Y_Z - 
     Z\wproj
     \|^2_2] 
     = & \sum_{j, e_{z,j} \in S^c}\underbrace{\sigma^2\sum_{i=1}^D\frac{\lambda_{z,j}^2}{\lambda_{x,i}^2}\langle e_{x,i}, e_{z,j}\rangle^2 \mathbbm{1}[\lambda_{x,i} > 0]}_{\Var_{z,j}} \nonumber
    + \sum_{j, e_{z,j} \in S}\underbrace{ \langle w^*, e_{z,j}\rangle^2 \lambda_{z,j}^2.\phantom{\Bigg|}}_{\Bias_{z,j}} 
\end{align} \label{eq:proj_bias}
\end{theorem}

The proof for this theorem is similar to the proof of Theorem \ref{thm:ols_loss} in that it uses the cyclic property of the trace to isolate the noise term. We then use the fact that each $e_{z,j} \in S$ is an eigenvector of $Z^\T Z$ to further decompose the expression. A full derivation of this decomposition is included in Appendix \ref{sec:appendix_bias_variance_decomp}. This case-like decomposition of the loss motivates our definition of the two different loss terms a single eigenvector $e_{z,j}$ can contribute to the overall expected loss. For a given eigenvector $e_{z,j}$ with associated eigenvalue $\lambda_{z,j}^2$, we will incur its \textbf{variance} loss if $e_{z,j} \not\in S$, and its \textbf{bias} loss if $e_{z,j} \in S$, where the variance loss $\Var_{z,j}$ and bias loss $\Bias_{z,j}$ are defined as:

\begin{equation}\label{eq:individual_loss_contributions}
     \Bias_{z,j} = \langle w^*, e_{z,j}\rangle^2 \lambda_{z,j}^2, \quad \Var_{z,j}  = \sigma^2\sum_{i=1}^D\frac{\lambda_{z,j}^2}{\lambda_{x,i}^2}\langle e_{x,i}, e_{z,j}\rangle^2\mathbbm{1}[\lambda_{x,i} > 0].
\end{equation}
$\Var_{z,j}$ is closely tied with the Spectral Inflation of an eigenvector, as $\Var_{z,j}$ will be large if $e_{z,j}$ demonstrates Spectral Inflation at test time. In this case if $e_{z,j} \not\in S$, $\wproj$ will have higher loss as a consequence of the label noise on the training examples distributed along this eigenvector. On the contrary, $\Bias_{z,j}$ is determined by the cosine similarity between the true labeling regressor $w^*$ and the eigenvector $e_{z,j}$. High cosine similarity means that this eigenvector makes a large contribution to determining a sample's label. If $e_{z,j} \in S$ and $e_{z,j}$ has a large cosine similarity to $w^*$, $\wproj$ will incur a high amount of loss as it is orthogonal to this important direction.

\subsection{Projection Reduces Out-of-Distribution Loss}

Thus far, we have presented a decomposition for the expected loss of an estimator that is equal to the pseudoinverse solution $\hat{w}$ projected into the ortho-complement of the span of the set $S \subseteq \Rows(V_Z^\top)$. In this subsection, we present a means for constructing the set $S$ to minimize the expected loss by comparing $\Var_{z,j}$ and $\Bias_{z,j}$ for each test eigenvector $e_{z,j}$.

The ideal set $S^* \subseteq \Rows(V_Z^\top)$ would consist solely of the eigenvectors $e_{z,j}$ that have a greater variance loss than bias loss. Formally, this set would be constructed using the following expression:
\begin{equation}\label{eq:best_eigenvector_set}
    S^* = \left\{e_{z,j}: e_{z,j} \in  \Rows(V_Z^\top), \Var_{z,j} \geq \Bias_{z,j} \right\}.
\end{equation}
The following theorem demonstrates that using the set $S^*$ would give us a regressor that achieves superior OOD performance than any other regressor produced using this projection procedure.

 \begin{theorem}\label{thm:best_projection_loss}
     Under the same assumptions as Theorem \ref{thm:ols_loss}, the regressor $\wproj^*$ constructed as in Equation \ref{eq:projected_regressor} using the set $S^*$ (cf. Equation \ref{eq:best_eigenvector_set}) can only improve on the OOD squared error loss of any other projected regressor $\wproj $ constructed as in Equation \ref{eq:projected_regressor} using a set $S \subseteq \Rows(V_Z^\top)$:
     \begin{align}
         \mathbb{E}[\|Y_Z - Z\wproj\|^2_2] \geq \mathbb{E}[\|Y_Z - Z\wproj^*\|^2_2].
     \end{align}
 \end{theorem}

 \begin{corollary}
      It follows from Theorem \ref{thm:best_projection_loss} and the fact that the pseudoinverse solution is a projected regressor using set $S = \{\}$ that:
     \begin{align}
         \mathbb{E}[\|Y_Z - Z\hat{w}\|^2_2] \geq \mathbb{E}[\|Y_Z - Z\wproj^*\|^2_2].
     \end{align}
 \end{corollary}

\subsection{Eigenvector Selection Under Uncertainty}\label{sec:eigselec_under_uncertainty}
Theorem \ref{thm:best_projection_loss} 
shows that a regressor based on the set $S^*$ 
works better OOD. Finding $S^*$ would be easy if we knew both $\Var_{z,j}$ and $\Bias_{z,j}$ for each test eigenvector $e_{z,j}$. While we can calculate $\Var_{z,j}$  directly, $\Bias_{z,j}$ requires the true weight vector $w^*$, and so we can only \textit{estimate} it using the pseudoinverse solution $\hat{w}$:

\begin{equation}
    \widehat{\Bias}_{z,j} = \langle\hat{w}, e_{z,j}\rangle^2\lambda_{z,j}^2 = (w^{*T}e_{z,j} + \epsilon^\T X^{\dagger \T}e_{z,j})^2\lambda_{z,j}^2.
    \label{eq:bias_hat}
\end{equation}

We fortunately have knowledge of some of the distributional properties of the dot product being squared: $\langle \hat{w}, e_{z,j}\rangle$.
In particular, $w^{*\T}e_{z,j}$ is a fixed but unknown scalar and $\epsilon^\T X^{\dagger \T}e_{z,j}$ is the linear combination of several i.i.d. Gaussian variables with zero mean and variance $\sigma^2$.
 \begin{equation}\label{eq:distribution_bias_hat}
    \epsilon^\T X^{\dagger \T}e_{z,j}\lambda_{z,j} \sim \mathcal{N}(0, \Var_{z,j}), \quad \langle \hat{w}, e_{z,j}\rangle \lambda_{z,j} \sim \mathcal{N}(\sqrt{\Bias_{z,j}}, \Var_{z,j}). 
 \end{equation}
The fact that $\widehat{\Bias}_{z,j}$ is a random variable makes it difficult to directly compare it with $\Var_{z,j}$. However, we can analyze the behavior of $\widehat{\Bias}_{z,j}$ when $\Bias_{z,j}$ is much larger than $\Var_{z,j}$, and vice versa, in order to devise a method for comparing these two quantities.

\noindent {\bf (Case 1): 
$\Bias_{z,j} \gg \Var_{z,j}$}.
In this case, $\Bias_{z,j} \approx \widehat{\Bias}_{z,j}$.
This is because $w^{*\T}e_{z,j}$ will be much greater than 
$\epsilon^\T X^{\dagger \T}e_{z,j}$, which causes the former term to dominate in the RHS of Equation~\ref{eq:bias_hat}.
Therefore $\widehat{\Bias}_{z,j} \gg \Var_{z,j}$.

\noindent {\bf (Case 2): $\Var_{z,j} \gg \Bias_{z,j}$}.
In this case, $\widehat{\Bias}_{z,j} \approx ( \epsilon^\T X^{\dagger \T}e_{z,j})^2\lambda_{z,j}^2$.
This is because $w^{*\T}e_{z,j}$ will be much smaller than 
$\epsilon^\T X^{\dagger \T}e_{z,j}$, which causes the latter term to dominate in the RHS of Equation~\ref{eq:bias_hat}. Therefore, since Equation \ref{eq:distribution_bias_hat} indicates $( \epsilon^\T X^{\dagger \T}e_{z,j})\lambda_{z,j}$ is a scalar Gaussian random variable, we know the distribution of its square:
\begin{equation}
    \widehat{\Bias}_{z,j} \sim \Var_{z,j}\times\chi^2_{df=1},
\end{equation}
where $\chi^2_{df=1}$ is a chi-squared random variable with one degree of freedom. If $\CDF^{-1}_{\chi^2_{df=1}}$ is the inverse CDF of the chi-squared random variable, then we have:
\begin{equation}
    \Pr(\widehat{\Bias}_{z,j} \leq \CDF^{-1}_{\chi^2_{df=1}} (\alpha)\times \Var_{z,j}) = \alpha.
\end{equation}

\noindent By applying these two cases, we can construct our set $S$ as follows:
\begin{equation}\label{eq:spar_set}
    S = \left\{  e_{z,j}: \widehat{\Bias}_{z,j} \leq \CDF^{-1}_{\chi^2_{df=1}} (\alpha)\times \Var_{z,j}\right\}.
\end{equation}
The intuition behind this case-by-case analysis is formalized with the following proposition and lemma:

\begin{proposition}\label{prop:probability}
Making the same assumptions as Theorem \ref{thm:ols_loss}, for a given choice of $\alpha \in [0,1]$, the probability that test eigenvector $e_{z,j}$ is included in our set $S$ as defined in \ref{eq:spar_set}:
\begin{equation}
    \Pr(e_{z,j} \in S) = 1 - Q_{\frac{1}{2}} \left(\sqrt{\frac{\Bias_{z,j}}{\Var_{z,j}}}, \sqrt{\CDF^{-1}_{\chi^2_{df=1}}(\alpha)} \right),
\end{equation}
where $Q_{\frac{1}{2}}$ is the Marcum Q-function with $M = \frac{1}{2}$.

\end{proposition}

\begin{lemma}\label{lem:tail_behaviour}
 Using the same assumptions as Proposition \ref{prop:probability}:
 \begin{equation}
     \Pr(e_{z,j} \in S) \xrightarrow{\frac{\Bias_{z,j}}{\Var_{z,j}} \to \infty} 0 , \quad\quad \Pr(e_{z,j} \in S) \xrightarrow{\frac{\Bias_{z,j}}{\Var_{z,j}} \to 0} \alpha.
 \end{equation}
\end{lemma}
Lemma \ref{lem:tail_behaviour} tells us that if we would incur significantly higher OOD loss from including $e_{z,j}$ in our set $S$ than excluding it, then $e_{z,j}$ \textbf{will not} be included in $S$.
Similarly, if we would incur significantly higher OOD loss from excluding $e_{z,j}$ in our set $S$ than including it, then $e_{z,j}$ \textbf{will} be included in $S$.

\subsection{Spectrally Adapted Regressor}

Creating $\wproj$ in this way yields \methodName, a regressor tailored for a specific covariate shift (see Algorithm~\ref{algo:main_algo}).
Finally, this procedure requires the the variance of the training label noise, $\sigma^2$.
We use a maximum likelihood estimate of this parameter~\citep{murphy2022probabilistic} from the training data.

\methodName takes as input a set of embedded train and test examples. Creating these representations is slightly less computationally expensive than simply performing inference on these two datasets, and far less computationally expensive than an additional training epoch and test set evaluation. \methodName also requires SVD to be performed on $X$ and $Z$, which is polynomial in the number of samples. Importantly, these additional computations only have to be performed a single time for each unique evaluation set. This is a stark contrast from other methods which require a computationally taxing regularizer to be computed with every batch \citep{ganin2016domain, sun_deep_2016, yao2022c}. Empirically, we find that using \methodName is much faster than other methods we compare with (Appendix \ref{sec:computational_cost}).

\begin{algorithm}[t]
\caption{\methodNameFull (\methodName)}\label{algo:main_algo}
{
\begin{algorithmic}
\Require Training Data $X, Y_X$, Unlabeled Test Distribution Data $Z$, Rejection Confidence $\alpha$
\State $\hat{w} \gets X^{\dagger}Y_X$ 
\State $U_X, D_X, V_X^\T \gets \mathrm{SVD}(X)$
\State $U_Z, D_Z, V_Z^\T \gets \mathrm{SVD}(Z)$
\State $\hat{\sigma}^2 \gets \mathrm{MLE}(X, Y_X)$ 
\State $S \gets \{\}$ \Comment{Initialize the set S as empty}

\For{$e_{z,j} \in \Rows(V_Z^\top), \lambda_{z,j} \in \mathrm{Diagonal}(D_Z)$ }
\Comment{Iterate over Z's singular vectors and values}
    \State  $\Var_{z,j} \gets \hat{\sigma^2} \sum_{i=1}^D\frac{\lambda_{z,j}^2}{\lambda_{x,i}^2}\langle e_{x,i}, e_{z,j}\rangle^2\mathbbm{1}[\lambda_{x,i} > 0]$
    \State  $\Bias_{z,j} \gets \langle \hat{w}, e_{z,j}\rangle^2 \lambda_{z,j}^2$
    \If{$(\CDF_{\chi^2}^{-1}(\alpha) \times \Var_{z,j}) \geq \Bias_{z,j}$}
        \State $S \gets S \cup \{e_{z_j}\}$ \Comment{Include this vector in $S$ if its bias is below its variance threshold}
\EndIf
\EndFor
\hspace{-7mm}\State $\wproj \gets \hat{w} - \sum_{e \in S}\langle \hat{w}, e\rangle e$  \Comment{Project out each of the selected vectors}

\hspace{-7mm}\Return $\wproj$
\end{algorithmic}
}
\end{algorithm}

\section{Experiments}\label{sec:experiments}

In this section, we apply \methodName to a suite of real-world and synthetic datasets to demonstrate its efficacy and explain how this method overcomes some shortcomings of the pseudoinverse solution.

Here we use models that are optimized using gradient-based procedures.
This contrasts with the main target of our analysis, the OLS solution (Equation \ref{eq:pseudoinv_soln}), as $\hat{w}$ is not found using an iterative procedure. Despite these differences, our analysis remains relevant as the optimality conditions of minimizing the squared error loss
ensure that gradient descent will converge to the OLS solution. 

\subsection{Synthetic Data}\label{sec:synthetic_data}
We establish a proof of concept by considering a synthetic data setting where we can carefully control the distribution shift under study.
Specifically, we apply our approach to two-dimensional Gaussian data following the data generative process described in Section \ref{sec:data_generative}.
Specifically, for experiments 1,2, and 3,  we sample our train and test data $X$ and $Z$ from origin-centered Gaussians with
diagonal covariance matrices, where the variances of $X$ and $Z$ are $(5, 10^{-5})$ and $(1, 40)$ respectively. For Experiment 4, there is no covariate shift and so the diagonal covariance matrices of $X$ and $Z$ are both $(1, 40)$.

We refer to the first and second indices of these vectors as the ``horizontal'' and ``vertical'' components and plot the vectors accordingly. In the first three experiments, the test distribution has much
more variance along the vertical component in comparison to the training distribution. We experiment with three different true labeling vectors:
$w^*_1 = (.01, .99999995)^T$; $w^*_2 = (0.9999995, 0.01)^T$; 
$w^*_3 = (\frac{1}{\sqrt{5}}, \frac{2}{\sqrt{5}})^T$.
The first two true labeling vectors represent functions that almost entirely depend on the vertical/horizontal component of the samples, respectively. $w_3^*$ depends on both directions, though it depends slightly more on the vertical component (cf. Figure \ref{fig:labels_vs_predictions}). For Experiment 4, we re-use $w_3^*$ as the true labelling vector as the focus of this experiment is the lack of covariate shift. For each labeling vector, we randomly sample $Z, X$ and $ \epsilon$ 10 times and calculate the squared error for three regression methods:
\begin{itemize}
    \item \textbf{OLS/Pseudoinverse Soution} (ERM): the minimizer for the training loss, $\hat{w} = X^{\dagger}Y_X$.
    \item \textbf{Principal Component Regression} \citep{bair2006prediction} (PCR): for this variant,  we calculate the OLS solution after projecting the data onto their first principal component.
    \item \textbf{ERM + \methodName}: the regressor produced by \methodName.
\end{itemize}

\begin{table*}[t!]
\centering
\resizebox{0.8\columnwidth}{!}{%
\begin{tabular}{ lcccc } 
  \hline
  \textbf{Synthetic Data} \\
  \hline
  Regression Method
  & Experiment 1 $(w^*_1)$ & Experiment 2 $(w^*_2)$  & Experiment 3 $(w^*_3)$ & Experiment 4 $(w^*_3)$\\   
  \hline
  \hline
 ERM
 & 2.54e6 $\pm$ 3.84e6 & 2.54e6 $\pm$ 3.84e6 & 2.54e6 $\pm$ 3.84e6 & \textbf{1.68e0} $\pm$  1.15e0\\  
  \hline
 PCR &  \textbf{1.59e5} $\pm$ 3.14e3 &  \textbf{1.12e0}$\pm$0.77e0 & \textbf{1.27e5} $\pm$ 2.51e3 & 8.04e2 $\pm$  1.30e1\\   
  \hline
 ERM + SpAR &  \textbf{1.59e5} $\pm$ 3.13e3 &  \textbf{2.81e0}$\pm$4.53e0 & \textbf{1.27e5} $\pm$ 2.51e3 & \textbf{1.68e0} $\pm$  1.15e0\\   \hline
\end{tabular}
}\caption{Mean and standard deviation of the squared error of our estimated regressors against various true labeling vectors.
Experiments 1, 2, and 3 use different true weight vectors, while Experiment 4 does not experience coviarate shift (see Section~\ref{sec:synthetic_data}).
}\label{tab:synthetic_results}
\end{table*}

We present the results of these experiments in Table \ref{tab:synthetic_results}. We first note that, for the first three experiments, $\hat{w}$ is expected to have the same error regardless of the true labeling vector. Second, $\wproj$ outperforms $\hat{w}$ regardless of which true regressor is chosen. Our projection method is most effective when $w^*_2$ is being used to label the examples. This is due to the fact that it relies mostly on the horizontal component of the examples, which has a similar amount of variance at both train and test time. As a result, \methodName is able to project out the vertical component while retaining the bulk of the true labeling vector's information. An example showing why this projection method is useful when $w^*_2$ is being used to label the examples is depicted in Figure \ref{fig:labels_vs_predictions}. Here, $\hat{w}$ significantly overestimates the influence of the vertical component on the samples' labels. \methodName is able to detect that it will not be able to effectively use the vertical component due to the large increase in variance as we move from train to test, and so it projects that component out of $\hat{w}$. Consequently, \methodName produces a labeling function nearly identical to the true labeling function.

PCR is able to achieve performance similar to \methodName on Experiments 1, 2, and 3. In these circumstances these methods are quite similar; these experiments have the second training principal component experiencing \textit{Spectral Inflation}, and so both methods achieve superior performance by projecting out the second principal component. In Experiment 4, however, no such \textit{Spectral Inflation} occurs, and so \methodName and ERM achieve performance far superior to PCR by leaving the OLS regressor intact. This demonstrates one of \methodName's most important properties: that it flexibly adapts to the covariate shift specified by the test data, rather than relying on the assumption that a certain adaptation will best perform OOD, as is the case with PCR.

\subsection{Tabular Datasets}\label{sec:tabular_data}

To test the efficacy of \methodName on real-world distribution shifts, we first experiment with two tabular datasets. Tabular data is 
common
in real-world machine learning applications and benchmarks,
particularly in the area of algorithmic fairness \citep{barocas-hardt-narayanan}. Therefore, it is important for any robust machine learning method to function well in this setting.

CommunitiesAndCrime, a popular dataset in fairness studies, provides a task where crime rates per capita must be predicted for different communities across the United States, with some states held out of the training data and used to form an OOD test set \citep{redmond2009communities,yao2022c}.
Skillcraft defines a task where one predicts the latency, in milliseconds, between professional video game players perceiving an action and making their own action \citep{blair2013skillcraft1}. An OOD test set is created by only including players from certain skill-based leagues in the train or test set.

We train neural networks with one hidden layer in the style of \citet{yao2022c}. We benchmark two methods:
\begin{itemize}
    \item \textbf{Standard Training (ERM)}: both the encoder and the regressor are trained in tandem to minimize the training objective using a gradient-based optimizer, in this case ADAM \citep{kingma2015adam}.
    \item \textbf{C-Mixup} \citep{yao2022c}: a data augmentation technique that generalizes the Mixup algorithm \citep{zhang2018mixup} to a regression setting. For this method, the encoder and regressor are  optimized to minimize the error on both the original samples and the synthetic examples produced by C-Mixup.
\end{itemize}
Data-augmentation techniques such as C-Mixup can be used in tandem with other techniques for domain adaptation, such as \methodName, to achieve greater results than either of the techniques on their own. Our results substantiate this.

We use the hyperparameters reported by \citet{yao2022c} when training both ERM and C-Mixup. After training, we apply \methodName to create a new regressor using the representations produced by the ERM model (ERM + \methodName) or C-Mixup model (C-Mixup + \methodName). 
For \methodName, we explored a few settings of the hyperparameter $\alpha$ (see Appendix
\ref{sec:alpha_sensitivity} for a discussion), and use a fixed value
of $\alpha = 0.999$ 
in all the experiments presented here.

These new regressors replace the learned regression weight in the last layer. We similarly benchmark the performance of the Pseudoinverse solution by replacing the last layer weight with $\hat{w}$ (ERM/C-Mixup + OLS).

Results from these tabular data experiments can be found in Figure \ref{fig:Tabular10Seeds}. Exact numbers are presented in Table \ref{tab:Tabular10Seeds} in the Appendix.

\begin{figure}[t!]
  \centering
  \includegraphics[width=0.9\linewidth]{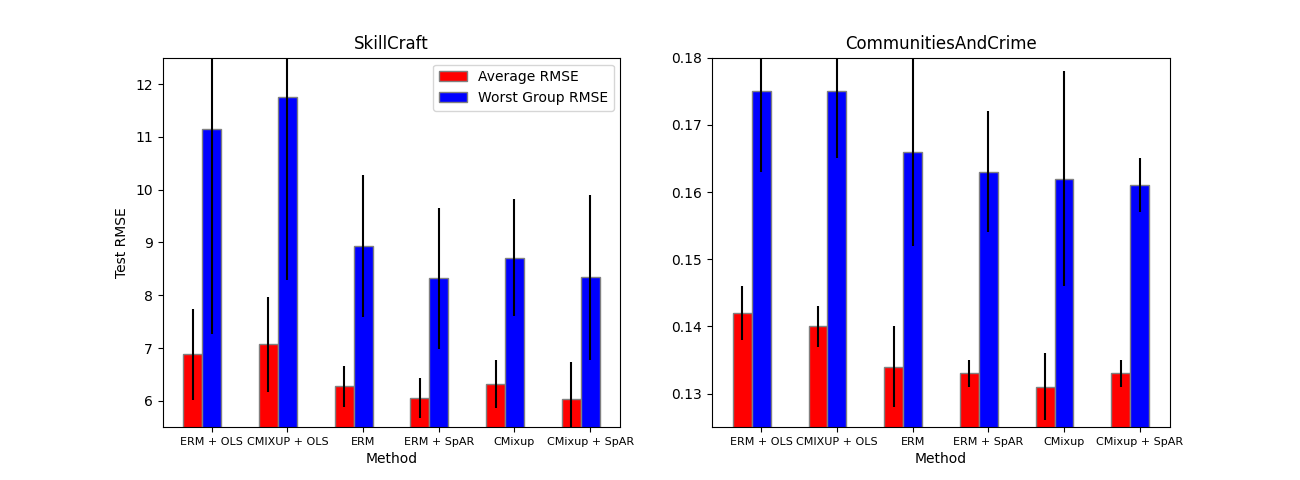}
\caption{\textbf{Tabular data.}
 OOD RMSE for several methods, each averaged across 10 seeds.
 }\label{fig:Tabular10Seeds}
  
\end{figure}

 Figure \ref{fig:Tabular10Seeds} shows that \methodName always produces a model with competitive or superior Average and Worst Group RMSE, regardless of the base model
that it is applied to. We also experiment with tuning the hyperparameters for both the ERM and C-Mixup models in Appendix \ref{app:tuned_baselines}.  With no additional tuning for \methodName specifically, \methodName yields a model with the strongest worst-group performance. 

\subsection{Image Datasets}

We now turn our attention to \methodName's efficacy on high-dimensional image datasets where a distribution shift is induced. Specifically, we experiment with the RCF-MNIST dataset from \citet{yao2022c}, as well as the ChairAngles-Tails dataset from \citet{gustafsson2023reliable}. RCF-MNIST tasks the model with predicting the angle of rotation for a series of images of clothing \citep{xiao2017fashion}. However, a spurious correlation between colour and rotation angle that is inverted at test time causes regressors focussing on this spurious feature to perform poorly when evaluated. The ChairAngles-Tails dataset requires the model to predict the angle of rotation for a synthetic image of a chair. A distribution shift is induced by only including certain rotation angles in the training set.

We benchmark ERM and C-Mixup, as well as two additional baseline methods:
\begin{itemize}
    \item \textbf{DANN} \citep{ganin2016domain}: in addition to minimizing the training loss, the encoder is trained to maximize the loss of an adversary trained to predict whether the representation comes from the training set or the test set.
    \item \textbf{Deep CORAL} \citep{sun_deep_2016}:  the encoder minimizes both the training loss and the difference between the first and second moments of the train and test data matrices.
\end{itemize}
 After performing a hyperparameter sweep (additional details in Appendix \ref{sec:appendix_hyperparameters}), we average results across 10 random seeds.  We apply SpAR to each of these baseline models using $\alpha = 0.999$ and no additional hyperparameter tuning. In the style of \citet{yao2022c}, we select the model checkpoint which best performed on a hold-out validation set as a form of early stopping. Results are presented in Table \ref{tab:Imagedatasets}.

 \methodName is regularly able to improve performance across a wide variety of architectures, tasks, and training methods. On RCF MNIST, \methodName makes improvements on ERM (0.6\% improvement), C-Mixup (1.2 \%improvement) , Deep CORAL (2\% improvement), and DANN (4\% improvement). On ChairAngles, \methodName makes improvements on ERM (0.5\% improvement), C-Mixup (0.8\% improvement) , Deep CORAL (2.3\% improvement), and DANN (1.6\% improvement).

The best performing baseline method varies across these two datasets, with ERM performing best on RCF MNIST and Deep CORAL performing best on ChairAngles-Tails. Despite these inconsistencies in baseline performance, \methodName consistently improves the performance of each method. This demonstrates \methodName's utility as a lightweight, efficient post processing method with a strong theoretical foundation that can be applied to a wide array of learned representations.

\begin{table*}[t!]
\centering
\resizebox{1.0\columnwidth}{!}{%
\begin{tabular}{ lcc } 
\hline
  \textbf{RCF-MNIST} \\
  \hline
Method & Baseline $(\downarrow)$ & Baseline + \methodName $(\downarrow)$\\
\hline
\hline
ERM & \textbf{0.155} $\pm$ 0.006 &\textbf{ 0.154} $\pm$ 0.006 \\
\hline
C-Mixup & 0.158 $\pm$ 0.011 & 0.156 $\pm$ 0.009 \\
\hline
Deep CORAL & 0.167 $\pm$ 0.012 & 0.165 $\pm$ 0.010\\
\hline
DANN & 0.177 $\pm$ 0.019 & 0.170 $\pm$ 0.015\\
\hline
\end{tabular}
\begin{tabular}{ lcc } 
\hline
  \textbf{ChairAngles-Tails} \\
  \hline
Method & Baseline $(\downarrow)$ & Baseline + \methodName $(\downarrow)$\\
\hline
\hline
ERM & 6.788 $\pm$ 0.634 & 6.753 $\pm$ 0.648 \\
\hline
C-Mixup & 6.504 $\pm$ 0.324 & 6.449 $\pm$ 0.325 \\
\hline
Deep CORAL & \textbf{5.978} $\pm$ 0.243 & \textbf{5.839} $\pm$ 0.259 \\
\hline
DANN & 6.440 $\pm$ 0.602 & 6.337 $\pm$ 0.603 \\
\hline
\end{tabular}
}
\caption{\textbf{Image data.}
OOD RMSE averaged across 10 seeds for models trained on RCF-MNIST and ChairAngles-Tails.
}
\label{tab:Imagedatasets}
\end{table*}

\subsection{PovertyMap - WILDS}\label{sec:povertymap}
We next examine the robustness of deep regression models under realistic distribution shifts in a high-dimensional setting. 
This experiment uses the PovertyMap-WILDS dataset~\citep{koh2021wilds}, where the task is to regress local satellite images onto a continuous target label representing an asset wealth index for the region.
PovertyMap provides an excellent test-bed for our method since, as seen in Figure \ref{fig:spectral_inflation}, DNNs attempting to generalize OOD on this dataset suffer from Spectral Inflation.

Once again, for ERM and C-Mixup we use the hyperparameters suggested by \citet{yao2022c} and for \methodName we use $\alpha = 0.999$. When training these baselines, we follow \citet{yao2022c} and select the model checkpoint which best performed on a hold-out validation set as a form of early stopping. These choices help to create strong baselines.
Results are presented in Table \ref{tab:Cmixpoverty}.

\begin{table*}[b!]
\centering
\resizebox{0.5\columnwidth}{!}{%
\begin{tabular}{ lcc } 
  \hline
Method & $\text{r}_{all} (\uparrow)$ & $\text{r}_{wg} (\uparrow)$\\
\hline
\hline
ERM & 0.793 $\pm$ 0.040 & 0.497 $\pm$ 0.099 \\
\hline
ERM + SpAR (Ours) & \textbf{0.794} $\pm$ 0.046 & \textbf{0.512} $\pm$ 0.092 \\
\hline
\hline
C-Mixup & 0.784 $\pm$ 0.045 & 0.489 $\pm$ 0.045 \\
\hline
C-Mixup + SpAR (Ours) & \textbf{0.794} $\pm$ 0.043 & \textbf{0.515} $\pm$ 0.091 \\
\hline
\end{tabular}
}
\caption{\textbf{PovertyMap-WILDS.} Average 
OOD
all-group and worst-group Spearman r across 5 splits.
}
\label{tab:Cmixpoverty}
\end{table*}

\begin{table*}[h!]
\centering
\resizebox{0.8\columnwidth}{!}{%
\begin{tabular}{ rlcc } 
  \hline
Robustness approach & Method & $\text{r}_{all} (\uparrow)$ & $\text{r}_{wg} (\uparrow)$\\
\hline
\hline
--- & ERM & \textbf{0.79} $\pm$ 0.04 & 0.50 $\pm$ 0.10 \\
\hline 
Data augmentation &  C-Mixup~\cite{yao2022c} & 0.78 $\pm$ 0.05 & 0.49 $\pm$ 0.05 \\
(pre-processing)  &  Noisy Student~\cite{xie_self-training_2020} & 0.76 $\pm$ 0.08 & 0.42 $\pm$ 0.11 \\
\hline 
Self-supervised pre-training &  SwAV~\cite{caron_unsupervised_2020} & 0.78 $\pm$ 0.06 & 0.45 $\pm$ 0.05 \\
(pre-processing)             &                                      &                 & \\
\hline 
Distribution alignment &  DANN~\cite{ganin2016domain} & 0.69 $\pm$ 0.04 & 0.33 $\pm$ 0.10 \\
(in-processing) &  DeepCORAL~\cite{sun_deep_2016} & 0.74 $\pm$ 0.05 & 0.36 $\pm$ 0.08 \\
                &  AFN~\cite{xu_larger_2019} & 0.75 $\pm$ 0.08 & 0.39 $\pm$ 0.08 \\
\hline
Subspace alignment &  RSD~\cite{chen_representation_2021} & 0.78 $\pm$ 0.03 & 0.44 $\pm$ 0.09 \\
(in-processing) &  DARE-GRAM~\cite{nejjar_dare-gram_2023} & 0.76 $\pm$ 0.06 & 0.44 $\pm$ 0.05 \\
\hline
 Spectral adaptation & \textbf{ERM + SpAR} (Ours) & \textbf{0.79} $\pm$ 0.04 & 0.51 $\pm$ 0.10 \\
 (post-processing)   &  \textbf{C-Mixup + SpAR} (Ours) &          \textbf{0.79} $\pm$ 0.04      &   \textbf{0.52} $\pm$ 0.08 \\                         
\hline
\hline
\end{tabular}
}
\caption{\textbf{PovertyMap-WILDS with unlabeled data.} In-processing methods and SpAR use unlabeled data that are distinct from the test set, but come from the same distribution~\citep{sagawa_extending_2022}.
}
\label{tab:povertymap_with_unlabeled_data}
\end{table*}

We can observe from Table \ref{tab:Cmixpoverty} that applying \methodName can significantly improve worst-group performance while maintaining competitive average performance. As with Section \ref{sec:tabular_data}, we further tune the hyperparameters for both the ERM and C-Mixup baselines in Appendix \ref{app:tuned_baselines}. 
With \textbf{no tuning of \methodName specifically}, it is able to enhance the tuned baseline and yield the strongest worst-group performance. \methodName is also more computationally efficient than other robustness methods (see Appendix \ref{sec:computational_cost}).

Additionally, we experiment with an unsupervised domain adaptation setting where we used unlabeled target domain data distinct from the test set to perform adaptation with \methodName \citep{sagawa_extending_2022}. We use the same base ERM and C-Mixup backbone models as presented in Table \ref{tab:Cmixpoverty}. We compare with many methods for robust ML, including some "in-processing" methods \citep{caron_unsupervised_2020} which use the unlabelled data to define an additional objective that is optimized during training. Results are presented in Table \ref{tab:povertymap_with_unlabeled_data}. We find that even when using a sample distinct from the evaluation data, the use of \methodName on either ERM or C-Mixup yields the best performance. The worst group performance of C-Mixup + \methodName achieves state of the art performance on the PovertyMap-WILDS leaderboard for methods using unlabeled target domain data \citep{sagawa_extending_2022}.

\section{Related Work}

Improving OOD performance is a critical and dynamic area of research.
Our approach follows in the tradition of Transductive Learning~\citep{gammerman2013learning} (adapting a model using unlabelled test data) and unsupervised Domain Adaptation~\citep{ben2006analysis,farahani2021brief} (using distributional assumptions to model train/test differences, then adapting using unlabeled test inputs).
Regularizing statistical moments between $P$ and $Q$ during training is a popular approach in unsupervised DA~\citep{gretton2009covariate} that has also been realized using deep neural networks~\citep{ganin2016domain,sun2017correlation}.
When transductive reasoning (adaptation to a test distribution) is not possible, additional structure in $P$---such as auxiliary labels indicating the ``domain'' or ``group'' that each training example belongs to---may be exploited to promote OOD generalization.
Noteworthy approaches include Domain Generalization~\citep{arjovsky2019invariant,gulrajani2020search} and Distributionally Robust Optimization~\citep{hu2018does, sagawa2019distributionally,levy2020large}.

Data augmentation is another promising avenue for improving OOD generalization~\citep{hendrycks2019benchmarking,ovadia2019can} that artificially increases the number and diversity of training set samples.
The recently proposed C-Mixup method 
focuses on regression under covariate shift;
it adapts the Mixup algorithm \citep{zhang2018mixup} to regression by upweighting the convex combination of training examples whose target values are similar. 
This pre-processing approach complements our post-processing adaptation approach; in our experiments we find that applying \methodName to a C-Mixup model often yields the best results.

In this work we investigate covariate shift in a regression setting by 
analyzing how the distribution shift affects eigenspectra of the source/target data.
We are not the first to study this problem, nor the first to use spectral properties in this investigation. \citet{pathak2022new} propose a new similarity measure between 
$P$ and $Q$ that can be used to bound the performance of non-parameteric regression methods under covariate shift.
\citet{wu2022power} analyze the sample efficiency of linear regression in terms of 
an eigendecomposition
of the second moment matrix of individual data points drawn from $P$ and $Q$.
Our work differs from these in that we go beyond an OOD theoretical analysis to propose a practical post-processing algorithm, which we find to be effective on real-world datasets.

\section{Conclusion}
This paper investigated the generalization properties of regression models when faced with covariate shift.
In this setting, our analysis shows that the Ordinary Least Squares solution---which minimizes the training risk---can fail dramatically OOD.
We attribute this sensitivity to \emph{Spectral Inflation}, where spectral subspaces with small variation during training see increased variation upon evaluation.
This motivates our adaptation method, \methodName, which uses unlabeled test data to estimate the subspaces with spectral inflation and project them away.
We apply our method to the last layer of deep neural regressors and find that it improves OOD performance on several synthetic and real-world datasets.

Our limitations include assumed access to unlabeled test data, and that the distribution shift in question is covariate shift.
Future work should focus on applying spectral adaptation to other distribution shifts (such as concept shift and subpopulation shift) and to the domain generalization setting.

\newpage

\bibliography{arxiv}

\newpage

\appendix

\section{OLS and the Pseudoinverse}\label{app:ols}

Classical statistics \citep{murphy2022probabilistic} tells us that when $X$ is full rank, the $\hat{w}$ minimizing this expression---known as the 
OLS
regressor---has the following form:

\begin{equation}
    \hat{w}_{OLS} = (X^\T X)^{-1}X^\T Y_X
\end{equation}

Of course, if $X$ is not full rank, the product $X^\T X$ 
cannot be inverted.
 In this case, the minimum norm solution can be constructed using the singular value decomposition of X. Specifically, X can be decomposed as $X = U_XD_XV_X^\T$. We can then construct the pseudoinverse of X by using $U_X, V_X$, and the matrix $D^{\dagger}$ which is given by taking the transpose of $D$, and replacing the diagonal singular value elements with their reciprocal.  In the case that the singular value is zero, the value of zero is used instead.
The pseudoinverse is then constructed as $X^{\dagger} = V_XD_X^{\dagger}U_X^\T$. Using these components, the minimum norm solution in the case of a degenerate X matrix is given by the following expression:

\begin{equation}
    \hat{w} = X^{\dagger}Y_X = V_XD_X^{\dagger}U_X^\T Y_X
\end{equation}

\section{Derivation of Loss of OLS Under Covariate Shift}\label{sec:appendix_ols_covariate}

We are interested in the following expression for the OOD risk of the OLS regressor:

\begin{equation}
    \mathrm{Risk}_{\mathrm{OOD}}(\hat w) = 
    \mathbb{E}[\|Y_Z - Z\hat{w}\|^2_2] = \mathbb{E}[\|Zw^* - ZX^{\dagger}Y\|^2_2] = \mathbb{E}[\|Zw^* - ZX^{\dagger}(Xw^* + \epsilon)\|^2_2]
\end{equation}

If we assume that $w^*$ exists within the span of the rows of X, then $X^{\dagger}X$ acts as an identity on $w^*$, giving us:
\begin{equation}
    = \mathbb{E}[\|ZX^{\dagger}\epsilon\|^2_2]
\end{equation}

The euclidean norm is $\|x\|_2 = \sqrt{x^\T x}$, so we can rephrase this expression as a scalar dot product. Scalars can be seen as $1 \times 1$ matrices, and are therefore equal to their trace. Therefore we can express this dot product as a trace in order to later use the cyclic property of the trace operator:

\begin{equation}
    = \mathbb{E}[\epsilon^\T X^{\dagger \T}Z^\T ZX^{\dagger}\epsilon] = \mathbb{E}[tr(\epsilon^\T X^{\dagger \T}Z^\T ZX^{\dagger}\epsilon)]
\end{equation}

We can cycle the trace and apply the properties of the trace of the product of two $N\times N$ matrices:

\begin{equation}
   =  \mathbb{E}[tr(\epsilon\epsilon^\T X^{\dagger \T}Z^\T ZX^{\dagger})] = \mathbb{E}[\sum^N_{i=1}\sum^N_{j=1}(\epsilon\epsilon^\T)_{i,j}(X^{\dagger \T}Z^\T ZX^{\dagger})_{i,j}]
\end{equation}

Since each entry of $\epsilon$ is independent from the other entries, and these entries follow the normal distribution $\mathcal{N}(0, \sigma^2)$, by applying the linearity of expectation we know that every term in this sum such that $i \neq j$ will be equal to zero, giving us:

\begin{equation}
    =  \sum^N_{i=1}\sum^N_{j=1}\mathbb{E}[(\epsilon\epsilon^\T)_{i,j}](X^{\dagger \T}Z^\T ZX^{\dagger})_{i,j} = \sum^N_{i=1}\sigma^2 (X^{\dagger \T}Z^\T ZX^{\dagger})_{i,i} = \sigma^2 tr(X^{\dagger \T}Z^\T ZX^{\dagger}) 
\end{equation}

We will use the singular value decompositions of these two matrices to simplify the expression further after cycling the trace:

\begin{equation}
= \sigma^2 tr(Z^\T ZX^{\dagger}X^{\dagger \T}) = \sigma^2 tr(V_ZD_Z^\T U_Z^\T U_ZD_ZV_Z^\T V_XD_X^{\dagger}U_X^\T U_XD_X^{\dagger \T}V_X^\T)
\end{equation}

\begin{equation}
     = \sigma^2 tr(V_ZD_Z^2V_Z^\T V_XD_X^{\dagger 2}V_X^\T) = \sigma^2 tr(D_X^{\dagger 2}V_X^\T
     V_ZD_Z^2V_Z^\T V_X)
\end{equation}

where $D_Z^2, D_X^{\dagger 2}$ are $D\times D$ diagonal matrices with diagonal values equal to the diagonal values of $D_Z$ and $D_X^{\dagger}$ squared, respectively. The $i^{th}$ diagonal entry of the matrix $V_X^\T V_ZD_Z^2V_Z^\T V_X$ is:

\begin{equation}
    \big[diag(V_X^\T V_ZD_Z^2V_Z^\T V_X)\big]_i = 
    \sum_{j=1}^D\lambda_{z,j}^2 \langle e_{x,i}, e_{z,j}\rangle^2
\end{equation}

Meaning that the entire expression will be equal to the value described:

\begin{equation}
    \mathrm{Risk}_{\mathrm{OOD}}(\hat w) =
    \sigma^2 \sum_{i=1}^D\sum_{j = 1}^D\frac{\lambda_{z,j}^2}{\lambda_{x,i}^2}\langle e_{x,i}, e_{z,j} \rangle^2\mathbbm{1}[\lambda_{x,i} > 0] .
\end{equation}

\section{Derivation of Bias-Variance Decomposition}\label{sec:appendix_bias_variance_decomp}

\methodName produces a regressor of the following form:
\begin{equation}
    \wproj = \hat{w} - \sum_{e \in S}\langle \hat{w}, e\rangle e
\end{equation}

Where we are projecting out a set of eignevectors $S$ from the pseudoinverse solution $\hat{w}$. We can substitute this into our expression for the OOD risk of a regressor to arrive at a bias-variance decomposition.

\begin{equation}
      \mathrm{Risk}_{\mathrm{OOD}}(\wproj) = \mathbb{E}[\|Zw^* - Z(\hat{w} - \sum_{e_{z,j} \in S}\langle \hat{w}, e_{z,j}\rangle e_{z,j}) \|^2_2]
\end{equation}
\begin{equation}
    = \mathbb{E}[\|-ZV_XD^{\dagger}_XU_X^\T \epsilon + Z\sum_{e_{z,j} \in S}(\epsilon^\T X^{\dagger\T}e_{z,j} + w^{*\T}e_{z,j})e_{z,j}\|^2_2]
\end{equation}
\begin{equation}
    = \mathbb{E}[\|-ZV_XD^{\dagger}_XU_X^\T \epsilon + Z\sum_{e_{z,j} \in S}\langle w^*, e_{z,j}\rangle e_{z,j} + Z\sum_{e_{z,j} \in S}\langle V_XD^{\dagger}_XU_X^\T\epsilon, e_{z,j}\rangle e_{z,j} \|^2_2]
\end{equation}

We can further simplify this expression by using the fact that the eigenvectors in $Rows(V_Z^\T)$ form an orthonormal basis, and so the sum of their outer products forms an identity matrix. Formally, $\sum_{j=1}^D e_{z,j}e_{z,j}^\T = I$. Using this on the leftmost term in the sum, we have:

\begin{equation}
    = \mathbb{E}[\|-Z\sum_{j=1}^D e_{z,j}e_{z,j}^\T V_XD^{\dagger}_XU_X^\T \epsilon + Z\sum_{e_{z,j} \in S}\langle w^*, e_{z,j}\rangle e_{z,j} + Z\sum_{e_{z,j} \in S}\langle V_XD^{\dagger}_XU_X^\T\epsilon, e_{z,j}\rangle e_{z,j}\|^2_2]
\end{equation}

\begin{equation}
    = \mathbb{E}[\|-Z\sum_{j=1}^D\langle V_XD^{\dagger}_XU_X^\T\epsilon, e_{z,j}\rangle e_{z,j} + Z\sum_{e_{z,j} \in S}\langle w^*, e_{z,j}\rangle e_{z,j} + Z\sum_{e_{z,j} \in S}\langle V_XD^{\dagger}_XU_X^\T\epsilon, e_{z,j}\rangle e_{z,j}\|^2_2]
\end{equation}
We can use the fact that $S \cup S^c$ form an orthogonal basis, where $S^c$ is the complement set of eigenvectors. We are also assuming that we are only projecting out vectors from the Z right singular vector basis. This gives us:
\begin{equation}
    \mathbb{E}[\|-Z\sum_{e_{z,j} \in S^c}\langle V_XD^{\dagger}_XU_X^\T\epsilon, e_{z,j}\rangle e_{z,j} + Z\sum_{e_{z,j} \in S}\langle w^*, e_{z,j}\rangle e_{z,j} \|^2_2] = \mathbb{E}[\|V - B\|^2_2]
\end{equation}

The euclidean norm $\|x\|_2 = \sqrt{x^\T x}$, and so we can consider the sum of products $V^\T V - 2V^\T B + B^\T B$. If we take the expectation over the error term $\epsilon$, which has mean 0, we are left with only $V^\T V + B^\T B$. 

$V^\T V$ is the error term we are already familiar with (Theorem \ref{thm:ols_loss}), restricted to the eigenvectors that weren't projected out:

\begin{equation}
    V^\T V = (Z\sum_{e_{z,j} \in S^c}\langle V_XD^{\dagger}_XU_X^\T\epsilon, e_{z,j}\rangle e_{z,j})^\T Z\sum_{e_{z,j} \in S^c}\langle V_XD^{\dagger}_XU_X^\T\epsilon, e_{z,j}\rangle e_{z,j} 
\end{equation}
\begin{equation}
    = (\sum_{e_{z,j} \in S^c}\langle V_XD^{\dagger}_XU_X^\T\epsilon, e_{z,j}\rangle e_{z,j})^\T Z^\T Z\sum_{e_{z,j} \in S^c}\langle V_XD^{\dagger}_XU_X^\T\epsilon, e_{z,j}\rangle e_{z,j}
\end{equation}

We note that each vector $e_{z,j} \in S^c$ is an eigenvector of $Z^\T Z $ with eigenvalue $\lambda_{z,j}^2$.

\begin{equation}
    = \sum_{e_{z,j} \in S^c}\langle V_XD^{\dagger}_XU_X^\T\epsilon, e_{z,j}\rangle e_{z,j}^\T \sum_{e_{z,j} \in S^c}\langle V_XD^{\dagger}_XU_X^\T\epsilon, e_{z,j}\rangle \lambda_{z,j}^2 e_{z,j}
\end{equation}

\begin{equation}
    = \sum_{e_{z,j}' \in S^c}\sum_{e_{z,j} \in S^c} \langle V_XD^{\dagger}_XU_X^\T\epsilon, e_{z,j}'\rangle \langle V_XD^{\dagger}_XU_X^\T\epsilon, e_{z,j}\rangle \lambda_{z,j}^2 e_{z,j}'^\T e_{z,j}
\end{equation}

Since $S^c$ is a subset of an orthonormal basis, we know that $e_{z,j}'^\T e_{z,j} = 1$ iff $e_{z,j}' = e_{z,j}$. Otherwise, $e_{z,j}'^\T e_{z,j} = 0$.

\begin{equation}
    = \sum_{e_{z,j} \in S^c}  \langle V_XD^{\dagger}_XU_X^\T\epsilon, e_{z,j}\rangle^2 \lambda_{z,j}^2 =  \sum_{e_{z,j} \in S^c} \epsilon^\T X^{\dagger\T} e_{z,j} e_{z,j}^T X^\dagger \epsilon \lambda_{z,j}^2
\end{equation}

In the expected loss, the expectation operator is applied to this expression, giving:

\begin{equation}
    \mathbb{E}[V^\T V]  = \mathbb{E}[  \sum_{e_{z,j} \in S^c} \epsilon^\T X^{\dagger\T} e_{z,j} e_{z,j}^T X^\dagger \epsilon \lambda_{z,j}^2]
\end{equation}

We can use the properties of the trace to isolate the label noise, as in Appendix \ref{sec:appendix_ols_covariate}:

\begin{equation}
    =  \sum_{e_{z,j} \in S^c} \sigma^2 tr(e_{z,j}^\T X^\dagger X^{\dagger \T}e_{z,j}) \lambda_{z,j}^2
\end{equation}

We can analyze the inner product of the vector $X^{\dagger \T}e_{z,j} = U_XD^{\dagger \T}_XV_X^\T e_{z,j}$ with itself:

\begin{equation}
    e_{z,j}^\T X^\dagger X^{\dagger \T}e_{z,j} = \sum_{i=1}^d \sum_{k=1}^d \frac{1}{\lambda_{x,i}} \langle e_{z,j}, e_{x,i} \rangle \frac{1}{\lambda_{x,k}} \langle e_{z,j}, e_{x,k} \rangle u_{x,i}^\T u_{x,k}\mathbbm{1}[\lambda_{x,i} > 0]\mathbbm{1}[\lambda_{x,k} > 0]
\end{equation}

Where $u_{x,i}$ is the $i^{th}$ column of $U_X$, i.e. the $i^{th}$ left singular vector of $X$. These left singular vectors also create an orthonormal basis, and so $u_{x,i}^\T u_{x,k} = 1$ iff $u_{x,i} = u_{x,k}$. Otherwise, $u_{x,i}^\T u_{x,k} = 0$. This ultimately gives us:

\begin{equation}
    \mathbb{E}[V^\T V]  = \sigma^2 \sum_{i=1}^D\sum_{j, e_{z,j} \in S^c}\frac{\lambda_{z,j}^2}{\lambda_{x,i}^2}\langle e_{x,i}, e_{z,j}\rangle^2\mathbbm{1}[\lambda_{x,i} > 0]
\end{equation}

We can use similar reasoning to show that bias term $B^\T B$ is a simple expression relying on the true weight vector:

\begin{equation}
    \mathbb{E}[B^T B] = B^T B = \sum_{e_{z,j} \in S}\langle w^*, e_{z,j}\rangle e_{z,j}^\T Z^\T Z\sum_{e_{z,j} \in S}\langle w^*, e_{z,j}\rangle e_{z,j}
\end{equation}
\begin{equation}
\end{equation}

\begin{equation}
     = \sum_{j, e_{z,j} \in S}\langle w^*, e_{z,j}\rangle^2 \lambda_{z,i}^2
\end{equation}

Therefore, we have the following expression for the expected loss:

\begin{equation}
    \mathbb{E}[\|Zw^* - Z(\hat{w} - \sum_{e_{z,j} \in S}\langle \hat{w}, e_{z,j}\rangle e_{z,j}) \|^2_2] = \mathbb{E}[V^\T V] + \mathbb{E}[B^\T B] 
\end{equation}

\begin{equation}
    =\sigma^2 \sum_{i=1}^D\sum_{j, e_{z,j} \in S^c}\frac{\lambda_{z,j}^2}{\lambda_{x,i}^2}\langle e_{x,i}, e_{z,j}\rangle^2\mathbbm{1}[\lambda_{x,i} > 0] + \sum_{j, e_{z,j} \in S}\langle w^*, e_{z,j}\rangle^2 \lambda_{z,j}^2
\end{equation}

\section{Proof of Theorem \ref{thm:best_projection_loss}}\label{sec:app_thm_3old}

In this section, we provide the proof of Theorem \ref{thm:best_projection_loss}.

This theorem compares the OOD squared error loss of two regressors, $\wproj$ and $\wproj^*$, which are constructed in the following way:
\begin{equation}
        \wproj = \hat{w} - \sum_{e \in S}\langle \hat{w}, e\rangle e, \quad \wproj^* = \hat{w} - \sum_{e \in S^*}\langle \hat{w}, e\rangle e
\end{equation}

We can invoke Theorem \ref{thm:bias_variance_decomp} to decompose the OOD squared error loss of the regressors:

\begin{equation}
    \mathbb{E}[\|Y_Z - Z\wproj\|^2_2]  = \sum_{e_{z,j} \in S^{c}} \Var_{z,j} + \sum_{e_{z,j} \in S} \Bias_{z,j}
\end{equation}
\begin{equation}
    \mathbb{E}[\|Y_Z - Z\wproj^*\|^2_2]  = \sum_{e_{z,j} \in S^{*c}} \Var_{z,j} + \sum_{e_{z,j} \in S^{*}} \Bias_{z,j}
\end{equation}

Since $\Rows(V_Z^\top) = S \cup S^c = S^{*} \cup S^{*c}$, we can decompose the losses into four sums:
\begin{equation}
    \mathbb{E}[\|Y_Z - Z\wproj\|^2_2]  = \sum_{e_{z,j} \in S^{c}\cap S^{*c}} \Var_{z,j} + \sum_{e_{z,j} \in S^{c}\cap S^{*}} \Var_{z,j} + \sum_{e_{z,j} \in S\cap S^*} \Bias_{z,j} + \sum_{e_{z,j} \in S\cap S^{*c}} \Bias_{z,j}.
\end{equation}
\begin{equation}
    \mathbb{E}[\|Y_Z - Z\wproj^*\|^2_2]  = \sum_{e_{z,j} \in S^{c}\cap S^{*c}} \Var_{z,j} + \sum_{e_{z,j} \in S^{c}\cap S^{*}} \Bias_{z,j} + \sum_{e_{z,j} \in S\cap S^*} \Bias_{z,j} + \sum_{e_{z,j} \in S\cap S^{*c}} \Var_{z,j}.
\end{equation}

This gives us:
\begin{equation}
    \mathbb{E}[\|Y_Z - Z\wproj\|^2_2] - \mathbb{E}[\|Y_Z - Z\wproj^*\|^2_2] = \sum_{e_{z,j} \in S^{c}\cap S^{*}}(\Var_{z,j} - \Bias_{z,j}) + \sum_{e_{z,j} \in S\cap S^{*c}}(\Bias_{z,j} -  \Var_{z,j}).
\end{equation}

By the definition of $S^*$, we know that $e_{z,j} \in S^*$ implies that $\Var_{z,j} \geq \Bias_{z,j}$. Therefore:
\begin{equation}
    \sum_{e_{z,j} \in S^{c}\cap S^{*}}(\Var_{z,j} - \Bias_{z,j}) \geq 0.
\end{equation}

Furthermore, for $e_{z,j} \not\in S^*$ and therefore in $S^{*c}$, it must be the case that $\Var_{z,j} < \Bias_{z,j}$. Therefore:
\begin{equation}
    \sum_{e_{z,j} \in S\cap S^{*c}}(\Bias_{z,j} -  \Var_{z,j}) \geq 0.
\end{equation}

This implies that the difference of OOD squared error losses is also greater or equal to zero, and therefore that $\wproj^*$ achieves superior loss.

\begin{equation}
    \mathbb{E}[\|Y_Z - Z\wproj\|^2_2] - \mathbb{E}[\|Y_Z - Z\wproj^*\|^2_2] = \sum_{e_{z,j} \in S^{c}\cap S^{*}}(\Var_{z,j} - \Bias_{z,j}) + \sum_{e_{z,j} \in S\cap S^{*c}}(\Bias_{z,j} -  \Var_{z,j}) \geq 0
\end{equation}

\begin{equation}
    \implies \mathbb{E}[\|Y_Z - Z\wproj\|^2_2] \geq \mathbb{E}[\|Y_Z - Z\wproj^*\|^2_2].
\end{equation}

\section{ Distribution of $\widehat{\Bias}$}

In Section \ref{sec:eigselec_under_uncertainty} we make statements about the distribution of $\widehat{\Bias}$. In this section, we further explain our reasoning for these claims.

\begin{equation}
    \widehat{\Bias}_{z,j} = \langle\hat{w}, e_{z,j}\rangle^2\lambda_{z,j}^2 = (w^{*T}e_{z,j} + \epsilon^\T X^{\dagger \T}e_{z,j})^2\lambda_{z,j}^2.
\end{equation}

We know that $\epsilon$ is a Gaussian vector with zero mean and spherical covariance. Therefore, $\epsilon^\T X^{\dagger \T}e_{z,j}\lambda_{z,j}$ would also have zero mean. For its covariance, we need only to multiply this expression by itself to recognize the expression from previous derivations:

\begin{equation}
    \mathbb{E}[e_{z,j}^\T X^\dagger \epsilon\epsilon^\T X^{\dagger \T}e_{z,j}\lambda_{z,j}^2]
\end{equation}

This expression is seen in the derivation of Theorem \ref{thm:bias_variance_decomp}, where we show it is equal to $\Var_{z,j}$. Therefore, the variance of $\epsilon^\T X^{\dagger \T}e_{z,j}\lambda_{z,j}$ is $\Var_{z,j}$. With this in mind, we can rewrite this expression as a scaling of a standard normal random variable:

\begin{equation}
    \epsilon^\T X^{\dagger \T}e_{z,j}\lambda_{z,j} = \sqrt{\Var_{z,j}}\beta, \quad \beta \sim \mathcal{N}(0, 1)
\end{equation}

We can also easily describe the distribution of $\langle \hat{w}, e_{z,j}\rangle \lambda_{z,j}$:

\begin{equation}
    \langle \hat{w}, e_{z,j}\rangle \lambda_{z,j} = w^{*T}e_{z,j}\lambda_{z,j} + \epsilon^\T X^{\dagger \T}e_{z,j}\lambda_{z,j}
\end{equation}

Which is a Gaussian random variable plus a constant, which shifts the mean of the Gaussian. This gives us the two distributions we list in Section \ref{sec:eigselec_under_uncertainty}:

 \begin{equation}\label{eqn:distributions_of_biashat}
    \epsilon^\T X^{\dagger \T}e_{z,j}\lambda_{z,j} \sim \mathcal{N}(0, \Var_{z,j}), \quad \langle \hat{w}, e_{z,j}\rangle \lambda_{z,j} \sim \mathcal{N}(\sqrt{\Bias_{z,j}}, \Var_{z,j}). 
 \end{equation}

 We would next like to explain the claims made in Case 2 of Section \ref{sec:eigselec_under_uncertainty}. Specifically, we make claims about the distribution of $\widehat{\Bias}_{z,j}$ when $\widehat{\Bias}_{z,j} \approx (\epsilon^\T X^{\dagger \T}e_{z,j})^2\lambda_{z,j}^2$:

 \begin{equation}
     \widehat{\Bias}_{z,j} \approx (\epsilon^\T X^{\dagger \T}e_{z,j})^2\lambda_{z,j}^2 = (\sqrt{\Var_{z,j}}\beta)^2 = \Var_{z,j}\beta^2 
 \end{equation}

 \begin{equation}
     \beta \sim \mathcal{N}(0,1), \quad \beta^2 \sim \chi^2(df=1)
 \end{equation}

 We therefore know in this case that $\widehat{\Bias}_{z,j}$ is the scaling of a chi-squared random variable. By properties of CDFs, we know that $\Pr(\Var_{z,j}\beta^2 \leq \alpha) = \Pr(\beta^2 \leq \frac{\alpha}{\Var_{z,j}})$, and therefore we know that the inverse CDF of $\Var_{z,j}\beta^2$ will be $\CDF^{-1}_{\chi^2_{df=1}} (\alpha)\times \Var_{z,j}$.

\section{Proof of Proposition 1}

First, we will restructure $\widehat{\Bias}_{z,j}$ as the scaling of a non-central chi-squared random variable. From Equation \ref{eqn:distributions_of_biashat}, we know the distribution of $\sqrt{\widehat{\Bias}_{z,j}}$, which we can write in terms of a Gaussian random variable with non-zero mean:

\begin{equation}
    \sqrt{\widehat{\Bias}_{z,j}} = \langle \hat{w}, e_{z,j}\rangle \lambda_{z,j} \sim \mathcal{N}(\sqrt{\Bias_{z,j}}, \Var_{z,j})
\end{equation}

\begin{equation}
    \implies \sqrt{\widehat{\Bias}_{z,j}} = \sqrt{\Var_{z,j}}\delta, \quad \delta \sim \mathcal{N}(\frac{\sqrt{\Bias}}{\sqrt{\Var_{z,j}}}, 1)
\end{equation}

We therefore know that $\delta^2$ is distributed according to a non-central chi-squared distribution:

\begin{equation}
    \widehat{\Bias}_{z,j} = (\sqrt{\Var_{z,j}}\delta)^2 = \Var_{z,j}\delta^2, \quad \delta^2 \sim \chi^2_{\lambda}(df=1, \lambda = \frac{\Bias_{z,j}}{\Var_{z,j}})
\end{equation}

Furthermore, we know the CDF of this variable as $\Pr(\Var_{z,j}\delta^2 \leq \alpha) = \Pr(\delta^2 \leq \frac{\alpha}{\Var_{z,j}})$.

We include an eigenvector $e_{z,j}$ in our set $S$ if $\widehat{\Bias}_{z,j} \leq \CDF^{-1}_{\chi^2_{df=1}} (\alpha)\times \Var_{z,j}$. The probability of this event occurring is given by the CDF of $\widehat{\Bias}_{z,j}$, which is the following:

\begin{equation}
    \Pr(\widehat{\Bias}_{z,j} \leq \CDF^{-1}_{\chi^2_{df=1}} (\alpha)\times \Var_{z,j}) = 1 - Q_{\frac{1}{2}}(\sqrt{\frac{\Bias_{z,j}}{\Var_{z,j}}}, \frac{\sqrt{\Var_{z,j}}\sqrt{\CDF^{-1}_{\chi^2_{df=1}}}  (\alpha)}{\sqrt{\Var_{z,j}}})
\end{equation}

\begin{equation}
    = 1 - Q_{\frac{1}{2}}(\sqrt{\frac{\Bias_{z,j}}{\Var_{z,j}}}, \sqrt{\CDF^{-1}_{\chi^2_{df=1}} (\alpha)})
\end{equation}

\section{Proof of Lemma \ref{lem:tail_behaviour}}

Proposition \ref{prop:probability} gives us an expression for the probability that a given eigenvector is included in the set $S$. Lemma \ref{lem:tail_behaviour} will use this proposition to demonstrate the tail behaviour of this expression. We will first note that since the expression in Proposition \ref{prop:probability} is a CDF, it is continuous. Therefore, in order to find its limits at 0 and $\infty$, we need only be able to evaluate the expression at these values. 

We will first show that:

\begin{equation}
     \Pr(e_{z,j} \in S) \xrightarrow{\frac{\Bias_{z,j}}{\Var_{z,j}} \to \infty} 0
\end{equation}

This is a special value of the Marcum Q function \citep{sun2008inequalities}. Specifically, $Q_{\frac{1}{2}}(\infty, b) = 1$ for any $b$. Therefore:

\begin{equation}
    \Pr(e_{z,j} \in S) = 1 - Q_{\frac{1}{2}}(\infty, \sqrt{\CDF^{-1}_{\chi^2_{df=1}} (\alpha)}) = 1-1 = 0
\end{equation}

We will next show that:

\begin{equation}
    \Pr(e_{z,j} \in S) \xrightarrow{\frac{\Bias_{z,j}}{\Var_{z,j}} \to 0} \alpha
\end{equation}

This is another special value of the Marcum Q function \citep{sun2008inequalities}. Specifically:

\begin{equation}
    Q_{\frac{1}{2}}(0, b) = \frac{\Gamma(\frac{1}{2}, \frac{b^2}{2})}{\Gamma(\frac{1}{2})}
\end{equation}

For any $b$. Here, $\Gamma$ with one argument is the gamma function and $\Gamma$ with two arguments is the upper incomplete gamma function. By properties of gamma functions, we know that if $\gamma$ is the lower incomplete gamma function, then $\Gamma(\frac{1}{2}, \frac{b^2}{2}) + \gamma(\frac{1}{2}, \frac{b^2}{2}) = \Gamma(\frac{1}{2})$. Using this property, and by letting $b = \sqrt{\CDF^{-1}_{\chi^2_{df=1}} (\alpha)}$, we have the following:

\begin{equation}
    \Pr(e_{z,j} \in S) =  1 - Q_{\frac{1}{2}}(0, b) = \frac{\Gamma(\frac{1}{2}, \frac{b^2}{2}) + \gamma(\frac{1}{2}, \frac{b^2}{2})}{\Gamma(\frac{1}{2})} - \frac{\Gamma(\frac{1}{2}, \frac{b^2}{2})}{\Gamma(\frac{1}{2})}
\end{equation}

\begin{equation}\label{eq:gamma_becomes_chi2}
    = \frac{\gamma(\frac{1}{2}, \frac{b^2}{2})}{\Gamma(\frac{1}{2})} = \CDF_{\chi^2_{df=1}} (\CDF^{-1}_{\chi^2_{df=1}} (\alpha)) = \alpha
\end{equation}

Where we have used the observation that the leftmost expression in Equation \ref{eq:gamma_becomes_chi2} is the CDF for a chi-squared distribution with one degree of freedom.

\section{Additional Training Details}

For our experiments in Section \ref{sec:experiments}, we adapt the code provided by \citet{yao2022c} in this Github repo: \url{https://github.com/huaxiuyao/C-Mixup}. While training, we perform early stopping on a validation set evaluation metric. For PovertyMap, this procedure is seen in the original work of  \citet{koh2021wilds}. We also use the hyperparameters provided in the appendix of \citet{yao2022c}'s work, including the following learning rates and bandwidth parameters for C-Mixup:

\begin{table*}[h!]
\centering
\resizebox{0.8\columnwidth}{!}{%
\begin{tabular}{ lccc } 
  \hline
  Hyperparameter
  & CommunitiesAndCrime & SkillCraft  & PovertyMap \\   
  \hline
  \hline
 Learning Rate
 & 1e-3 & 1e-2 & 1e-3\\  
  \hline
 Bandwidth &  1.0 &  5e-4 & 0.5\\  

  \hline
\end{tabular}
}
\caption{Hyperparameters used for training models responsible for the results in Section \ref{sec:experiments}.}
\label{tab:base_hyperparams}
\end{table*}

We additionally make the modification to train models without a bias term in the final linear layer. This is due to the fact that \methodName assumes a regressor that does not use a bias.

Models are trained using Tesla T4 GPUs from NVIDIA. Tabular and synthetic experiments take less than 10 minutes to run for a single seed and hyperparameter setting. PovertyMap experiments take roughly 3 hours to run when training ERM and roughly 15 hours to run when training C-Mixup.

\section{Tabular Data Results with Base Hyperparameters}

In this section, we provide the table of results that Figure \ref{fig:Tabular10Seeds} is based upon. This is the performance of the models using the hyperparameters described in Table \ref{tab:base_hyperparams}. The results are included in Table \ref{tab:Tabular10Seeds}.

\begin{table*}[t!]
\centering
\resizebox{1.0\columnwidth}{!}{%
\begin{tabular}{ lcc } 
\hline
  \textbf{SkillCraft} \\
  \hline
Method & Average RMSE $(\downarrow)$ & Worst Group RMSE $(\downarrow)$\\
\hline
\hline
ERM & 6.273 $\pm$ 0.384 & 8.933 $\pm$ 1.338 \\
\hline
ERM + OLS & 6.884 $\pm$ 0.860 & 11.156 $\pm$ 3.892 \\
\hline
ERM + \methodName (Ours) & \textbf{6.049} $\pm$ 0.379 & \textbf{8.317} $\pm$ 1.327 \\
\hline
\hline
C-Mixup & 6.319 $\pm$ 0.450 & 8.713 $\pm$ 1.106 \\
\hline
C-Mixup + OLS & 7.070 $\pm$ 0.898 & 11.747 $\pm$ 3.450 \\
\hline
C-Mixup + \methodName (Ours) & \textbf{6.038} $\pm$ 0.705 & \textbf{8.343} $\pm$ 1.563 \\
\hline
\end{tabular}
\begin{tabular}{ lcc } 
\hline
  \textbf{CommunitiesAndCrime} \\
  \hline
Method & Average RMSE $(\downarrow)$& Worst Group RMSE $(\downarrow)$\\
\hline
\hline
ERM & 0.134 $\pm$ 0.006 & 0.166 $\pm$ 0.014 \\
\hline
ERM + OLS & 0.142 $\pm$ 0.004 & 0.175 $\pm$ 0.012 \\
\hline
ERM + \methodName (Ours) & \textbf{0.133} $\pm$ 0.002 & \textbf{0.163} $\pm$ 0.009 \\
\hline
\hline
C-Mixup & \textbf{0.131} $\pm$ 0.005 & 0.162 $\pm$ 0.016 \\
\hline
C-Mixup + OLS & 0.140 $\pm$ 0.003 & 0.175 $\pm$ 0.010 \\
\hline
C-Mixup + \methodName (Ours) & 0.133 $\pm$ 0.002 & \textbf{0.161} $\pm$ 0.004 \\
\hline
\end{tabular}
}
\caption{\textbf{Tabular data.}
OOD RMSE averaged across 10 seeds for models using the hyperparameters described in Table \ref{tab:base_hyperparams}.
}
\label{tab:Tabular10Seeds}
\end{table*}

\section{Hyperparameter Search}\label{sec:appendix_hyperparameters}

For hyperparameter tuning, we perform random search over the learning rate and the bandwidth used in C-Mixup. Specifically, we search over learning rates using the following formula for the learning rate $lr$ and bandwidth $bw$:

\begin{equation}
    lr = base_{lr}*10^{u}, u \sim Unif(-1, 1)
\end{equation}
\begin{equation}
    bw = base_{bw}*10^{u}, u \sim Unif(-1, 1)
\end{equation}
where $base_{lr}$ and $base_{bw}$ are the values described in Table \ref{tab:base_hyperparams} for each dataset. We test out 10 randomly selected hyperparameter settings for both ERM and C-Mixup, and select the settings that yield the best validation performance.  Those hyperparameter settings selected for C-Mixup are presented in Table \ref{tab:cmix_tuned_hyperparams} and hyperparameter settings selected for ERM are presented in Table \ref{tab:erm_tuned_hyperparams}.

\begin{table*}[h!]
\centering
\resizebox{0.8\columnwidth}{!}{%
\begin{tabular}{ lccc } 
  \hline
  Hyperparameter
  & CommunitiesAndCrime & SkillCraft  & PovertyMap \\   
  \hline
  \hline
 Learning Rate
 & 0.003630376073213171 & 0.023276939100527687 & 0.003630376073213171\\  
  \hline
 Bandwidth &  0.35090148857968506 &  0.0013316008334250096 & 0.17545074428984253\\  

  \hline
\end{tabular}
}
\caption{Tuned hyperparameters used for training C-Mixup models. \ref{sec:experiments}.}
\label{tab:cmix_tuned_hyperparams}
\end{table*}

\begin{table*}[h!]
\centering
\resizebox{0.8\columnwidth}{!}{%
\begin{tabular}{ lccc } 
  \hline
  Hyperparameter
  & CommunitiesAndCrime & SkillCraft  & PovertyMap \\   
  \hline
  \hline
 Learning Rate
 & 0.008246671732726021 & 0.023276939100527687 & 0.003630376073213171\\  
  \hline
\end{tabular}
}
\caption{Tuned hyperparameters used for training ERM models.}
\label{tab:erm_tuned_hyperparams}
\end{table*}

For RCF-MNIST and ChairAngles-Tails, we use a similar procedure, only instead using a base learning rate of $7e-5$ for RCF-MNIST and $0.001$ for ChairAngles-Tails. For the bandwidth, we use a base bandwidth of $0.2$ for RCF-MNIST and $5e-4$ for ChairAngles-Tails. Additionally, Deep CORAL and DANN require a penalty weight $pw$ which we generate using the same procedure and a base $pw$ of $1.0$. These hyperparameters are presented in Tables \ref{tab:rcfmnist_tuned_hyperparams} and \ref{tab:chairangles_tuned_hyperparams}.

\begin{equation}
    pw = base_{pw}*10^{u}, u \sim Unif(-1, 1)
\end{equation}

\begin{table*}[h!]
\centering
\resizebox{0.8\columnwidth}{!}{%
\begin{tabular}{ lccc } 
  \hline
  Method/Hyperparameter
  & Learning Rate & Bandwidth  & Penalty Weight \\   
  \hline
  \hline
 ERM
 & 3.851230830192189e-05 & - & -\\  
  \hline
 C-Mixup &  3.636910217027964e-05 &  0.20091864782463165  & -\\  
  \hline
 Deep CORAL &  5.2547676552479794e-05 & - & 3.721492665736836 \\  
  \hline
 DANN &  7.100098449013041e-05  & - & 0.10654593849857387\\  
  \hline
\end{tabular}
}
\caption{Tuned hyperparameters used for training models on RCF-MNIST.}
\label{tab:rcfmnist_tuned_hyperparams}
\end{table*}

\begin{table*}[h!]
\centering
\resizebox{0.8\columnwidth}{!}{%
\begin{tabular}{ lccc } 
  \hline
  Method/Hyperparameter
  & Learning Rate & Bandwidth  & Penalty Weight \\   
  \hline
  \hline
 ERM
 & 0.0003572140318996373 & - & -\\  
  \hline
 C-Mixup &  0.0007506810936068544  &  0.0018607463328684177  & -\\  
  \hline
 Deep CORAL &  0.008246671732726021 & - & 5.647617424572879 \\  
  \hline
 DANN & 0.0010142997784304342  & - & 0.10654593849857387 \\  
  \hline
\end{tabular}
}
\caption{Tuned hyperparameters used for training models on ChairAngles-Tails.}
\label{tab:chairangles_tuned_hyperparams}
\end{table*}

\section{Tuned Baselines}\label{app:tuned_baselines}

Using the hyperparameters presented in Tables \ref{tab:cmix_tuned_hyperparams} and \ref{tab:erm_tuned_hyperparams} which were selected hyperparameter search process described in Section \ref{sec:appendix_hyperparameters}, we benchmark the performance of ERM and C-Mixup models across 10 seeds for the tabular datasets and the 5 data folds for PovertyMap.  We report results for PovertyMap and the tabular datasets in Tables \ref{tab:tuned_poverty}  and \ref{tab:tuned_tabular}, respectively.

We find that \methodName can achieve superior worst group performance than any other method presented in either Tables \ref{tab:tuned_poverty} or \ref{tab:tuned_tabular}, or in Section \ref{sec:experiments}. For C-Mixup on CommunitiesAndCrime, we see that tuning hyperparameters on the validation set yields poorer performance (Table \ref{tab:tuned_tabular}) than using the hyperparameters presented in \citet{yao2022c}'s work (Table \ref{tab:Tabular10Seeds}). However, we can see that a \methodName model is able to achieve the best worst-group RMSE of any model on this dataset, 0.161.

\begin{table*}[t!]
\centering
\resizebox{1.0\columnwidth}{!}{%
\begin{tabular}{ lcc } 
\hline
  \textbf{SkillCraft} \\
  \hline
Method & Average RMSE $(\downarrow)$ & Worst Group RMSE $(\downarrow)$\\
\hline
\hline
ERM & \textbf{5.917} $\pm$ 0.620 & 8.308 $\pm$ 1.915 \\
\hline
ERM + OLS & 6.548 $\pm$ 0.915 & 10.219 $\pm$ 3.123 \\
\hline
ERM + \methodName (Ours) & 6.083 $\pm$ 0.681 & \textbf{8.193} $\pm$ 1.212 \\
\hline
\hline
C-Mixup & \textbf{5.816} $\pm$ 0.558 & 8.371 $\pm$ 1.611 \\
\hline
C-Mixup + OLS & 6.535 $\pm$ 0.822 & 10.297 $\pm$ 2.362 \\
\hline
C-Mixup + \methodName (Ours) & 5.833 $\pm$ 0.580 & \textbf{7.922} $\pm$ 1.043 \\
\hline
\end{tabular}
\begin{tabular}{ lcc } 
\hline
  \textbf{CommunitiesAndCrime} \\
  \hline
Method & Average RMSE $(\downarrow)$& Worst Group RMSE $(\downarrow)$\\
\hline
\hline
ERM & \textbf{0.133} $\pm$ 0.004 & \textbf{0.161} $\pm$ 0.010 \\
\hline
ERM + OLS & 0.149 $\pm$ 0.018 & 0.184 $\pm$ 0.032 \\
\hline
ERM + \methodName (Ours) & 0.134 $\pm$ 0.007 & 0.164 $\pm$ 0.013 \\
\hline
\hline
C-Mixup & 0.133 $\pm$ 0.003 & 0.171 $\pm$ 0.012 \\
\hline
C-Mixup + OLS & 0.144 $\pm$ 0.011 & 0.177 $\pm$ 0.019 \\
\hline
C-Mixup + \methodName (Ours) & \textbf{0.132} $\pm$ 0.004 & \textbf{0.164} $\pm$ 0.008 \\
\hline
\end{tabular}
}
\caption{\textbf{Tabular data.}
OOD RMSE averaged across 10 seeds for models using tuned hyperparameters.
}
\label{tab:tuned_tabular}
\end{table*}

\begin{table*}[bt!]
\centering
\resizebox{0.5\columnwidth}{!}{%
\begin{tabular}{ lcc } 
  \hline
Method & $\text{r}_{all} (\uparrow)$ & $\text{r}_{wg} (\uparrow)$\\
\hline
\hline
ERM & 0.798 $\pm$ 0.052 & 0.518 $\pm$ 0.076 \\
\hline
ERM + \methodName (Ours) & \textbf{0.799} $\pm$ 0.045 & \textbf{0.522} $\pm$ 0.080 \\
\hline
\hline
C-Mixup & \textbf{0.806} $\pm$ 0.031 & 0.523 $\pm$ 0.083 \\
\hline
C-Mixup + \methodName (Ours) & 0.803 $\pm$ 0.038 & \textbf{0.528} $\pm$ 0.087 \\
\hline
\end{tabular}
}
\caption{\textbf{PovertyMap-WILDS.} Average 
OOD
all-group and worst-group Spearman r across 5 splits for models using tuned hyperparameters.
}
\label{tab:tuned_poverty}
\end{table*}

\section{Sensitivity of Alpha Hyperparameter}\label{sec:alpha_sensitivity}

\begin{figure}[h!]
  \centering
  \includegraphics[width=0.85\linewidth]{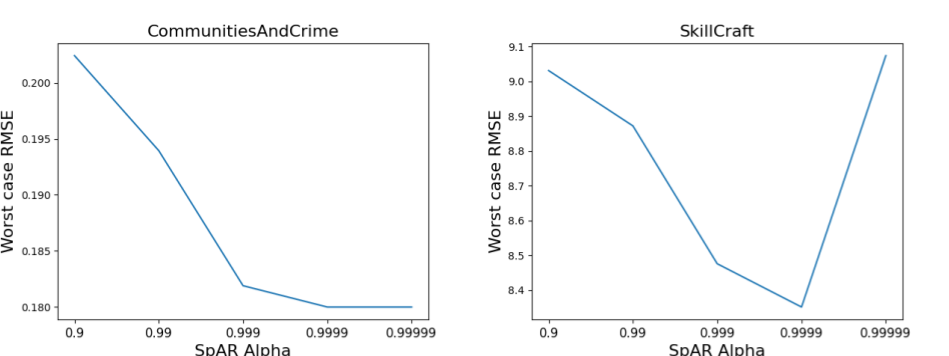}
  \caption{\textbf{Hyperparameter sensitivity} SpAR performance as a function of $\alpha$ on tabular datasets.
  }\label{fig:spar_alpha_search}
  
\end{figure}

Throughout this work, we use a single setting of $\alpha$ for each of our experiments. Our specific setting of $\alpha$=0.999 was selected using a minimal amount of tuning on a single seed of a single experiment. This value was then used on every seed of every dataset, regardless of potential improvements. To achieve a more complete understanding of SpAR’s sensitivity to $\alpha$, we conduct an experiment measuring OOD performance as a function of $\alpha$ when SpAR is applied to an ERM base model on the SkillCraft and CommunitiesAndCrime datasets. See Figure \ref{fig:spar_alpha_search} for results. We see that on the CommunitiesAndCrime dataset, a higher $\alpha$ than 0.999 could have resulted in superior worst case performance. Meanwhile, on SkillCraft, we clearly see that setting $\alpha$ too close to 1 can result in very poor worst group performance. Expression \ref{eq:spar_set} in the paper indicates that as $\alpha$ tends towards zero, the regressor produced by SpAR will more closely resemble the solution produced by OLS. Specifically, fewer eigenvectors will be projected out from the OLS solution. Conversely, as $\alpha$ tends towards one, the regressor produced by SpAR will tend towards the zero vector. This can be seen as a tradeoff between the cases where no Spectral Inflation is expected and where Spectral Inflation is expected to occur along every right singular vector.

In general, selecting $\alpha$ using validation set performance can have mixed results, as SpAR is intended to produce a regressor for a specific evaluation set (namely, the OOD test set, not the ID validation set). Future work could investigate the interesting question of how $\alpha$ could be selected based on the amount of spectral inflation presented in the train/evaluation data.

\section{Computational Cost}\label{sec:computational_cost}

The computational cost of SpAR comes from collecting the representations (running forward passes for every train and test example) and performing SVD, with the former step dominating the cost. Notably, it is much less cumbersome than other adaptation techniques. Computing the SVD of the matrix can be done in polynomial time, and we find in practice that performing this one-time post-hoc adaptation is quite efficient relative to other methods that must compute a regularizer or augment data on each training iteration (see Table \ref{tab:timing_povery_map}).

\begin{table*}[h!]
\centering
\begin{tabular}{ lc } 
  \hline
Method & Average RMSE \\
\hline
ERM & 3h22m $\pm$ 0h22m \\
\hline
C-Mixup & 14h58 $\pm$ 1h01m \\
\hline
DARE-GRAM & 5h58m $\pm$ 0h26m \\
\hline
ERM + SpAR (Ours) & 4h11m $\pm$ 0h33m \\
\hline
SpAR only & 0h40m $\pm$ 0h18m \\
\hline
\end{tabular}
\caption{\textbf{Measured train time} on PovertyMap. Each model is trained on a NVIDIA Tesla T4 GPU. In-processing methods and SpAR use a large pool of unlabeled data that are distinct from the test set, but come from the same distribution~\cite{sagawa_extending_2022}.}
\label{tab:timing_povery_map}
\end{table*}

\section{Limitations and Broader Impacts}
\methodName is designed for covariate shift, and its ability to handle other types of distribution shift (such as concept shift) is not known analytically.
To be more specific, we assume that the targets have a the same linear relationship (via the ground truth weight $w^*$) with inputs $X$ and $Z$, and that $X$ and $Z$ are covariate-shifted.
A subtle issue here is that when $X$ and $Z$ are internal representations of some neural net, we require that the difference $P$ and $Q$ is captured in terms of a covariate shift \emph{in the representation space}, which may or may not correspond to a covariate shift in the original input space (which could be some high-dimensional vector, e.g. pixels).

Empirically, however, we successfully apply \methodName to several real-world datasets without assurance that they exhibit only covariate shift, and find promising results.
The spectral inflation property that we observe in real data (Figure \ref{fig:spectral_inflation}) may be relevant to other distribution shifts as well, although this remains to be seen in future studies.
Identifying covariate shift within a datasets is an active area of work~\citep{ginsberg2022learning} that complements our efforts in this paper.

Our research seeks to improve OOD generalization with the hopes of ensuring ML benefits are distributed more equitably across social strata.
However, it is worthwhile to be self-reflexive about the methodology we use when working towards this goal.
For example, for the purposes of comparing against existing methods from the literature, we use the Communities and Crime dataset, where average crime rates are predicted based on statistics of neighborhoods, which could include demographic information.
This raises a potential fairness concern: even if we have an OOD-robust model, it may not be fair if it uses demographic information in its predictions.
While this is not the focus of our paper, we note that the research community is in the process of reevaluating tabular datasets used for benchmarking~\citep{ding2021retiring,bao2021s}.

\end{document}